\documentclass[10pt,journal,compsoc]{IEEEtran}

\usepackage{graphicx}
\usepackage{hyperref}
\usepackage{multirow}
\usepackage{booktabs}
\usepackage{float}
\usepackage{placeins}
\usepackage{bbm}
\usepackage{dsfont}
\usepackage{eufrak}
\usepackage{color}

\ifCLASSOPTIONcompsoc
  \usepackage[nocompress]{cite}
\else
  \usepackage{cite}
\fi
\ifCLASSINFOpdf
\else
\fi
\usepackage{bm}
\usepackage{amsmath}
\usepackage{amsfonts}

\makeatletter
\newcommand{\printfnsymbol}[1]{%
  \textsuperscript{\@fnsymbol{#1}}%
}
\makeatother

\begin{document}
\title{Dynamic Neural Networks: A Survey}
\author{Yizeng~Han\IEEEauthorrefmark{1},
        Gao~Huang\IEEEauthorrefmark{1},~\IEEEmembership{Member,~IEEE,}
        Shiji~Song,~\IEEEmembership{Senior~Member,~IEEE,}
        Le~Yang,
        Honghui~Wang,
        and~Yulin~Wang
\IEEEcompsocitemizethanks{\IEEEcompsocthanksitem Yizeng Han, Gao Huang, Shiji Song, Le Yang, Honghui Wang and Yulin Wang are with the Department of Automation, Tsinghua University, Beijing 100084, China. Gao Huang is also with Beijing Academy of Artificial Intelligence, Beijing, 100084. E-mail: \{hanyz18, yangle15, wanghh20, wang-yl19\}@mails.tsinghua.edu.cn; \{gaohuang, shijis\}@tsinghua.edu.cn. Corresponding author: Gao Huang.
}
}


\IEEEtitleabstractindextext{%
\vskip -0.15in
\begin{abstract}

    Dynamic neural network is an emerging research topic in deep learning. Compared to static models which have fixed computational graphs and parameters at the inference stage, dynamic networks can adapt their structures or parameters to different inputs, leading to notable advantages in terms of accuracy, computational efficiency, adaptiveness, etc. In this survey, we comprehensively review this rapidly developing area by dividing dynamic networks into three main categories: 1) {\emph{sample-wise}} dynamic models that process each sample with data-dependent architectures or parameters; 2) \emph{spatial-wise} dynamic networks that conduct adaptive computation with respect to different spatial locations of image data; and 3) \emph{temporal-wise} dynamic models that perform adaptive inference along the temporal dimension for sequential data such as videos and texts. The important research problems of dynamic networks, e.g., architecture design, decision making scheme, optimization technique and applications, are reviewed systematically. Finally, we discuss the open problems in this field together with interesting future research directions.
\end{abstract}
\vskip -0.1in
\begin{IEEEkeywords}
Dynamic networks, Adaptive inference, Efficient inference, Convolutional neural networks.
\end{IEEEkeywords}}
\maketitle
\begingroup\renewcommand\thefootnote{\IEEEauthorrefmark{1}}
\footnotetext{\emph{Equal contribution.}}
\endgroup

\IEEEdisplaynontitleabstractindextext
\IEEEpeerreviewmaketitle

\vspace{-2ex}
\IEEEraisesectionheading{\section{Introduction}\label{sec:introduction}}
\IEEEPARstart{D}{eep} neural networks (DNNs) are playing an important role in various areas including computer vision (CV) \cite{alexnet,Simonyan15,szegedy2015going,he2016deep,huang2017densely} and natural language processing (NLP) \cite{vaswani2017attention,devlin_bert_2019,brown2020language}. 
Recent years have witnessed many successful deep models such as AlexNet \cite{alexnet}, VGG \cite{Simonyan15}, GoogleNet \cite{szegedy2015going}, ResNet \cite{he2016deep}, DenseNet \cite{huang2017densely} and Transformer \cite{vaswani2017attention}. These {architectural innovations} have enabled the training of deeper, more accurate and more efficient models. The recent research on neural architecture search (NAS) \cite{zoph2016neural, liu_darts_2018} further speeds up the process of designing more powerful structures. However, most of the prevalent deep learning models perform inference in a static manner, i.e., both the computational graph and the network parameters are fixed once trained, which may limit their representation power, efficiency and interpretability \cite{graves2016adaptive,huang2017multi,yang2019condconv,sabour2017dynamic}. 

Dynamic networks, as opposed to static ones, can adapt their structures or parameters to the input during inference, and therefore enjoy favorable properties that are absent in static models. In general, dynamic computation in the context of deep learning has the following advantages:

\begin{figure*}
  \centering
  \vspace{-2ex}
    \includegraphics[width=0.875\linewidth]{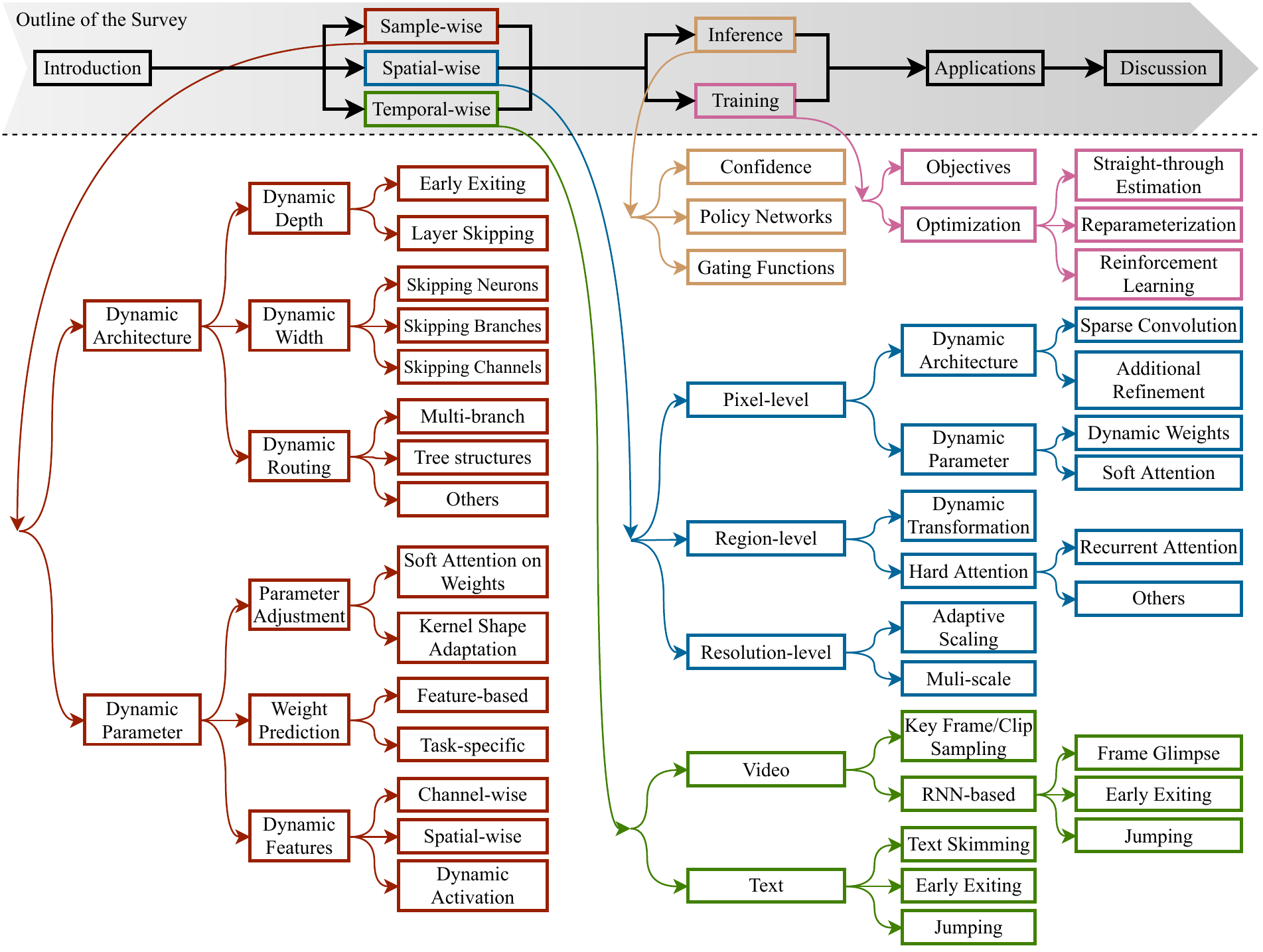}
    \vskip -0.175in
    \centering\caption{Overview of the survey. {We first review the dynamic networks that perform adaptive computation at three different granularities (i.e. sample-wise, spatial-wise and temporal-wise). Then we summarize the decision making strategy, training technique and applications of dynamic models. Existing open problems in this field together with some future research directions are finally discussed. Best viewed in color.}}
    \label{framework}
    \vspace{-3ex}
\end{figure*}

\textbf{1) Efficiency.} One of the most notable advantages of dynamic networks is that they are able to allocate computations on demand at test time, by selectively activating model components (e.g. layers \cite{huang2017multi}, channels \cite{lin2017runtime} or sub-networks \cite{shazeer2017outrageously}) \emph{conditioned} on the input. Consequently, less computation is spent on canonical samples that are relatively easy to recognize, or on less informative spatial/temporal locations of an input. {In addition to \emph{computational efficiency}, dynamic models have also shown promising results for improving \emph{data efficiency} in the scenario of few-shot learning \cite{bertinetto2016learning,wang2019tafe}.}


\textbf{2) Representation power.} Due to the data-dependent network architecture/parameters, dynamic networks have significantly enlarged parameter space and improved representation power. For example, with a minor increase of computation, model capacity can be boosted by applying feature-conditioned attention weights on an ensemble of convolutional kernels \cite{yang2019condconv,chen_dynamic_2020_attentionOver}. It is worth noting that the popular soft attention mechanism could also be unified in the framework of dynamic networks, as different channels \cite{hu2018squeeze}, spatial areas \cite{woo_cbam_2018} or temporal locations \cite{yang_neural_2017} of features are dynamically re-weighted at test time. 

\textbf{3) Adaptiveness.} Dynamic models are able to achieve a desired trade-off between accuracy and efficiency for dealing with varying computational budgets on the fly. Therefore, they are more adaptable to different hardware platforms and changing environments, compared to static models with a fixed computational cost.

\textbf{4) Compatibility.} Dynamic networks are compatible with most advanced techniques in deep learning, including architecture design \cite{he2016deep,huang2017densely}, optimization algorithms \cite{kingma2014adam,DBLP:journals/corr/IoffeS15} and data preprocessing \cite{wang2019implicit,DBLP:journals/corr/abs-1805-09501}, which ensures that they can benefit from the most recent advances in the field to achieve state-of-the-art performance. For example, dynamic networks can inherit architectural innovations in lightweight models \cite{howard2017mobilenets}, or be designed via NAS approaches \cite{zoph2016neural, liu_darts_2018}. Their efficiency could also be further improved by acceleration methods developed for static models, such as network pruning \cite{huang2018condensenet}, weight quantization \cite{hubara2016binarized}, knowledge distillation \cite{hinton2014distilling} and low-rank approximation \cite{low_rank}.

\textbf{5) Generality.} As a substitute for static deep learning techniques, many dynamic models are general approaches that can be applied seamlessly to a wide range of applications, such as image classification \cite{huang2017multi, yang_resolution_2020}, object detection \cite{figurnov2017spatially} and semantic segmentation \cite{li_not_2017}. Moreover, the techniques developed in CV tasks are proven to transfer well to language models in NLP tasks \cite{dehghani_universal_2019, elbayad_depth-adaptive_2020}, and vice versa.

\textbf{6) Interpretability.} We finally note that the research on dynamic networks {may bridge} the gap between the underlying mechanism of deep models and brains, as it is believed that the brains process information in a dynamic way~\cite{hubel1962receptive,murata2000selectivity}. It is possible to analyze which components of a dynamic model are activated \cite{yang_resolution_2020} when processing an input sample, and to observe which parts of the input are accountable for certain predictions \cite{wang2020glance}. These properties may shed light on interpreting the decision process of DNNs.



\begin{table}
  \vspace{-1ex}
  \caption{Notations used in this paper.}
  \label{tab:notations}
  \vspace{-4ex}
  \begin{center}
    \begin{tabular}{c|c}
      \hline
      \textbf{Notations} & \textbf{Descriptions} \\
      \hline
      $\mathbb{R}^m$ & $m$-dimensional real number domain \\
      \hline
      $a, \mathbf{a}$ & Scalar, vector/matrix/tensor \\
      \hline
      $\mathbf{x,y}$ & Input, output feature \\
      \hline
      $\mathbf{x}^{\ell}$ & Feature at layer $\ell$ \\
      \hline
      $\mathbf{h}_t$ & Hidden state at time step $t$ \\
      \hline
      $\mathbf{x(p)}$ & Feature at spatial location $\mathbf{p}$ on $\mathbf{x}$ \\
      \hline
      $\bm{\Theta}$ & Learnable parameter \\
      \hline
      $\bm{\hat{\Theta}}|\mathbf{x}$ & Dynamic parameter conditioned on $\mathbf{x}$ \\
      \hline
      $\mathbf{x}\star\mathbf{W}$ & Convolution of feature $\mathbf{x}$ and weight $\mathbf{W}$ \\
      \hline
      $\otimes$ & Channel-wise or element-wise multiplication \\
      \hline
      $\mathcal{F}(\cdot,\bm{\Theta})$ & Functional Operation parameterized by $\bm{\Theta}$ \\
      \hline
      $\mathcal{F}\circ\mathcal{G}$ & Composition of function $\mathcal{F}$ and $\mathcal{G}$ \\
      \hline
    \end{tabular}
  \end{center}
  \vskip -0.35in
\end{table}

In fact, adaptive inference, the key idea underlying dynamic networks, has been studied before the popularity of modern DNNs. The most classical approach is building a model ensemble through a cascaded \cite{viola_robust_2004} or parallel \cite{jacobs1991adaptive} structure, and selectively activating the models conditioned on the input. Spiking neural networks (SNNs) \cite{maass1997networks,izhikevich2003simple} also perform data-dependent inference by propagating pulse signals. However, the training strategy for SNNs is quite different from that of popular convolutional neural networks (CNNs), and they are less used in vision tasks. Therefore, we leave out the work related to SNNs in this survey. 

\begin{figure}
  \centering
    \vspace{-1ex}
    \includegraphics[width=\linewidth]{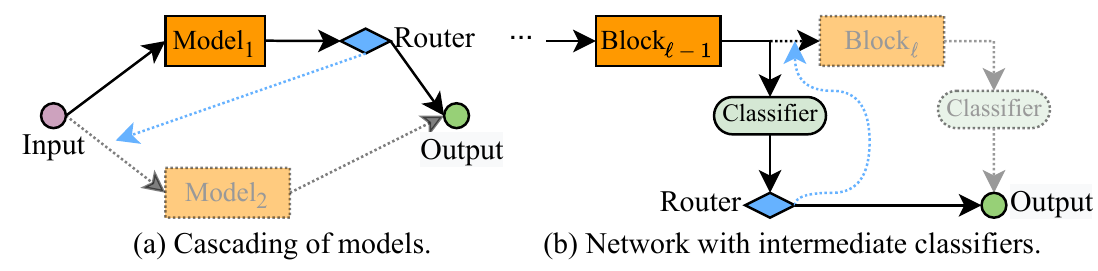}
    \vskip -0.15in
    \caption{Two early-exiting schemes. The dashed lines and shaded modules are not executed, conditioned on the decisions made by the routers.}
    \label{cascading_models}
    \vskip -0.2in
\end{figure}

\begin{figure*}
  \centering
  \vspace{-2ex}
    \includegraphics[width=\linewidth]{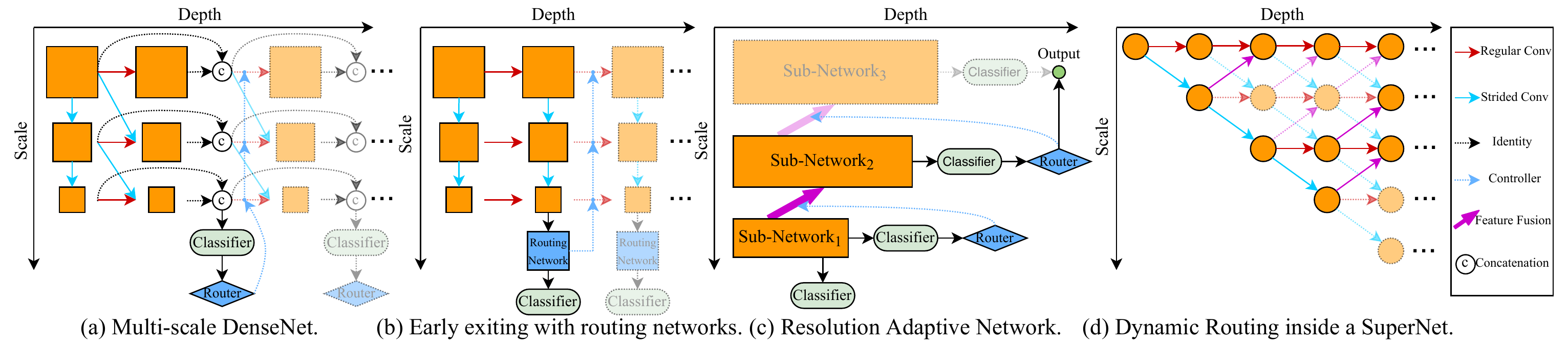}
    \vskip -0.175in
    \caption{{Multi-scale architectures with dynamic inference graphs. The first three models (a, b, c) perform adaptive early exiting with specific architecture designs and exiting policies. Dynamic routing is achieved inside a SuperNet (d) to activate data-dependent inference paths.}}
    \label{multi_scale}
    \vspace{-3ex}
\end{figure*}
In the context of deep learning, dynamic inference with modern deep architectures, has raised many new research questions and has attracted great research interests in the past three years.
Despite the extensive work on designing various types of dynamic networks, a systematic and comprehensive review on this topic is still lacking. This motivates us to write this survey, to review the recent advances in this rapidly developing area, with the purposes of 1) providing an overview as well as new perspectives for researchers who are interested in this topic; 2) pointing out the close relations of different subareas and reducing the risk of reinventing the wheel; and 3) summarizing the key challenges and possible future research directions.

{This survey is organized as follows (see \figurename~\ref{framework} for an overview). In Sec. \ref{sec_sample_wise}, we introduce the most common \emph{sample-wise} dynamic networks which adapt their architectures or parameters conditioned on each input sample. Dynamic models working on a finer granularity, i.e., \emph{spatially} adaptive and \emph{temporally} adaptive models, are reviewed in Sec. \ref{sec_spatially_adaptive} and Sec.\ref{sec_temporal_adaptive}, respectively\footnote{{These two categories can also be viewed as sample-wise dynamic networks as they perform adaptive computation within each sample at a finer granularity, and we adopt such a split for narrative convenience.}}. Then we investigate the decision making strategies and the training techniques of dynamic networks in Sec. \ref{inference_and_train}. The applications of dynamic models are further summarized in Sec. \ref{sec_tasks}. Finally, we conclude this survey with a discussion on a number of open problems and future research directions in Sec. \ref{sec:discussion}. For better readability, we list the notations that will be used in this survey in Table \ref{tab:notations}.}

\vspace{-2ex}
\section{{Sample-wise Dynamic Networks}}
\label{sec_sample_wise}
Aiming at processing different inputs in data-dependent manners, sample-wise dynamic networks are typically designed from two perspectives: 1) adjusting model architectures to allocate appropriate computation based on each sample, and therefore reducing redundant computation for increased efficiency (Sec. \ref{dynamic_arch}); 
2) adapting network \emph{parameters} to every input sample with fixed computational graphs, with the goal of boosting the representation power with minimal increase of computational cost (Sec. \ref{adaptive_params}).

\vspace{-2ex}
\subsection{Dynamic Architectures}
\label{dynamic_arch}
Considering different inputs may have diverse computational demands, it is natural to perform inference with dynamic architectures conditioned on each sample. Specifically, one can adjust the network depth (Sec. \ref{dynamic_depth}), width (Sec. \ref{dynamic_width}), or perform dynamic routing within a super network (SuperNet) that includes multiple possible paths (Sec. \ref{SuperNets}). Networks with dynamic architectures not only save redundant computation for canonical ("easy") samples, but also preserve their representation power when recognizing non-canonical ("hard") samples. Such a property leads to remarkable advantages in efficiency compared to the acceleration techniques for static models \cite{huang2018condensenet,liu2017learning, hubara2016binarized}, which handle "easy" and "hard" inputs with identical computation, and fail to reduce intrinsic computational redundancy.

\vspace{-1.5ex}
\subsubsection{Dynamic Depth} \label{dynamic_depth}
As modern DNNs are getting increasingly deep for recognizing more "hard" samples, a straightforward solution to reducing redundant computation is performing inference with dynamic depth, which can be realized by 1) \emph{early exiting}, i.e. allowing "easy" samples to be output at shallow exits without executing deeper layers \cite{teerapittayanon2016branchynet,bolukbasi2017adaptive,huang2017multi}; or 2) \emph{layer skipping}, i.e. selectively skipping intermediate layers conditioned on each sample \cite{graves2016adaptive,wang2018skipnet,veit2018convolutional}.

\begin{figure*}
  \vspace{-2ex}
  \centering
    \includegraphics[width=\linewidth]{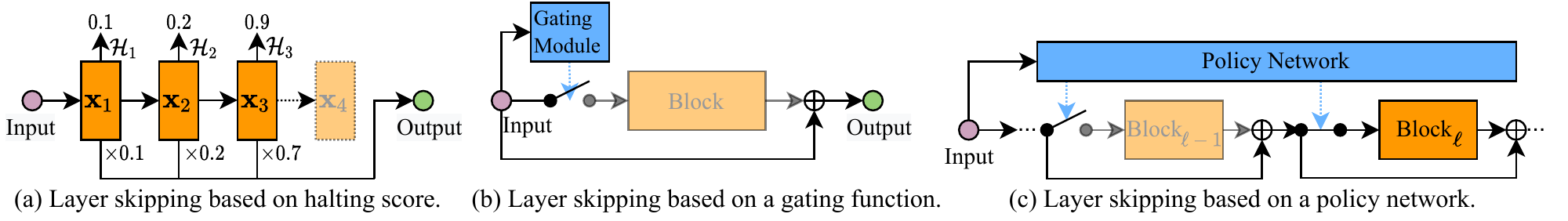}
    \vskip -0.175in
    \caption{Dynamic layer skipping. Feature $\mathbf{x}_4$ in (a) are not calculated conditioned on the halting score, and the gating module in (b) decides whether to execute the block based on the intermediate feature. The policy network in (c) generates the skipping decisions for all layers in the main network.}
    \label{skipping_layers}
    \vspace{-3ex}
\end{figure*}
\noindent\textbf{1) Early exiting.} The complexity (or "difficulty") of input samples varies in most real-world scenarios, and shallow networks are capable of correctly identifying many canonical inputs. Ideally, these samples should be output at certain early exits without executing deeper layers.

For an input sample $\mathbf{x}$, the forward propagation of an $L$-layer deep network $\mathcal{F}$ could be represented by 
\begin{equation}
  \setlength{\abovedisplayskip}{3pt}
  \mathbf{y} = \mathcal{F}^{L}\circ\mathcal{F}^{L-1}\circ\cdots\circ\mathcal{F}^1(\mathbf{x}),
  \setlength{\belowdisplayskip}{3pt}
\end{equation}
\noindent where $\mathcal{F}^{\ell}$ denotes the operational function at layer $\ell,1\!\le \ell\!\le L$. In contrast, early exiting allows to terminate the inference procedure at an intermediate layer. For the $i$-th input sample $\mathbf{x}_i$, the forward propagation can be written as
\begin{equation}
  \setlength{\abovedisplayskip}{3pt}
  \mathbf{y}_i = \mathcal{F}^{\ell_i}\circ\mathcal{F}^{\ell_i-1}\circ\cdots\circ\mathcal{F}^1(\mathbf{x}_i), 1\!\le \ell_i\!\le L.
  \setlength{\belowdisplayskip}{3pt}
\end{equation}
Note that $\ell_i$ is adaptively determined based on $\mathbf{x}_i$. Extensive architectures have been studied to endow DNNs with such early exiting behaviors, as discussed in the following.


a) \emph{Cascading of DNNs.} The most intuitive approach to enabling early exiting is cascading multiple models (see \figurename~\ref{cascading_models} (a)), and adaptively retrieving the prediction of an early network without activating latter ones. For example, Big/little-Net \cite{park2015big} cascades two CNNs with different depths. After obtaining the \emph{SoftMax} output of the first model, early exiting is conducted when the score margin between the two largest elements exceeds a threshold. {Moreover, a number of classic CNNs \cite{alexnet,szegedy2015going,he2016deep} are cascaded in \cite{bolukbasi2017adaptive} and \cite{wang2017idk}}. After each model, a decision function is trained to determine whether the obtained feature should be fed to a linear classifier for immediate prediction, or be sent to subsequent models.

b) \emph{Intermediate classifiers.}
The models in the aforementioned cascading structures are mutually independent. Consequently, 
once a "difficult" sample is decided to be fed to a latter network, a whole inference procedure needs to be executed from scratch without reusing the already learned features. A more compact design is involving intermediate classifiers within one backbone network (see \figurename~\ref{cascading_models} (b)), so that early features can be propagated to deep layers if needed. Based on such a  multi-exit architecture, adaptive early exiting is typically achieved according to confidence-based criteria \cite{teerapittayanon2016branchynet,leroux2017cascading} or learned functions \cite{bolukbasi2017adaptive,guan2018energy,dai_epnet_2020}.


c) \emph{Multi-scale architecture with early exits.} Researchers \cite{huang2017multi} have observed that in chain-structured networks, the multiple classifiers may interfere with each other, which degrades the overall performance. A reasonable interpretation could be that in regular CNNs, the high-resolution features lack the coarse-level information that is essential for classification, leading to unsatisfying results for early exits. Moreover, early classifiers would force the shallow layers to generate \emph{task-specialized} features, while a part of \emph{general} information is lost, resulting in degraded performance for deep exits. To tackle this issue, multi-scale dense network (MSDNet) \cite{huang2017multi} adopts 1) a \emph{multi-scale} architecture, {which consists of multiple sub-networks for processing feature maps with different resolutions (scales)}, to quickly generate coarse-level features that are suitable for classification; 2) \emph{dense connections}, to reuse early features and improve the performance of deep classifiers (see \figurename~\ref{multi_scale} (a)). Such a specially-designed architecture effectively enhances the overall accuracy of all the classifiers in the network. 

{Based on the multi-scale architecture design, researchers have also studied the exiting policies \cite{mcgill2017deciding, jie2019anytime} (see \figurename~\ref{multi_scale} (b)) and training schemes \cite{li2019improved} of early-exiting dynamic models. More discussion about the inference and training schemes for dynamic models will be presented in Sec. \ref{inference_and_train}.}


Previous methods typically achieve the adaptation of network depths. From the perspective of exploiting spatial redundancy in features, resolution adaptive network (RANet, see \figurename~\ref{multi_scale} (c)) \cite{yang_resolution_2020} first processes each sample with low-resolution features, while high-resolution representations are conditionally utilized based on early predictions.


Adaptive early exiting is also extended to language models (e.g. BERT \cite{devlin_bert_2019}) for improving their efficiency on NLP tasks \cite{liu_fastbert_2020,xin_deebert_2020,schwartz_right_2020,zhou_bert_2020}. In addition, it can be implemented in recurrent neural networks (RNNs) for \emph{temporally} dynamic inference when processing sequential data such as videos \cite{fan_watching_2018,wu_dynamic_2020} and texts \cite{shen_reasonet_2017,yu_fast_2018,liu_finding_2020} (see Sec. \ref{sec_temporal_adaptive}).

\noindent\textbf{2) Layer skipping.} The general idea of the aforementioned early-exiting paradigm is skipping the execution of all the deep layers after a certain classifier. More flexibly, the network depth can also be adapted on the fly by strategically skipping the calculation of \emph{intermediate layers} without placing extra classifiers. Given the $i$-th input sample $\mathbf{x}_i$, dynamic layer skipping could be generally written as
\begin{equation}
  \setlength{\abovedisplayskip}{3pt}
  \mathbf{y}_i = (\mathds{1}^{L}\circ\mathcal{F}^{L})\circ (\mathds{1}^{L-1}\circ\mathcal{F}^{L-1})\circ\cdots\circ (\mathds{1}^1\circ\mathcal{F}^1)(\mathbf{x}_i),
  \setlength{\belowdisplayskip}{3pt}
\end{equation}
where $\mathds{1}^{\ell}$ denotes the indicator function determining the execution of layer $\mathcal{F}^{\ell}, 1\!\le\!\ell\!\le\!L$.
This scheme is typically implemented on structures with skip connections (e.g. ResNet \cite{he2016deep}) to guarantee the continuity of forward propagation, and here we summarize three common approaches.

a) \emph{Halting score} is first proposed in \cite{graves2016adaptive}, where an accumulated scalar named halting score adaptively decides whether the hidden state of an RNN will be directly fed to the next time step. The halting scheme is extended to vision tasks by viewing residual blocks within a ResNet stage \footnote{Here we refer to a stage as a stack of multiple residual blocks with the same feature resolution.} as linear layers within a step of RNN \cite{figurnov2017spatially} (see \figurename~\ref{skipping_layers} (a)).
Rather than skipping the execution of layers with independent parameters, multiple blocks in each ResNet stage could be replaced by one weight-sharing block, leading to a significant reduction of parameters \cite{leroux_iamnn_2018}. In every stage, the block is executed for an adaptive number of steps according to the halting score.

In addition to RNNs and CNNs, the halting scheme is further implemented on Transformers \cite{vaswani2017attention} by \cite{dehghani_universal_2019} and \cite{elbayad_depth-adaptive_2020} to achieve dynamic network depth on NLP tasks.


b) {\emph{Gating function} is also a prevalent option for dynamic layer skipping due to its plug-and-play property.} Take ResNet as an example (see \figurename~\ref{skipping_layers} (b)), let $\mathbf{x}^{\ell}$ denote the input feature of the $\ell$-th residual block, gating function $\mathcal{G}^{\ell}$ generates a binary value to decide the execution of residual block $\mathcal{F}^{\ell}$. This procedure could be represented by\footnote{For simplicity and without generality, the subscript for sample index will be omitted in the following.}
\begin{equation}
  \setlength{\abovedisplayskip}{3pt}
  \mathbf{x}^{\ell+1} = \mathcal{G}^{\ell}(\mathbf{x}^{\ell})\mathcal{F}^{\ell}(\mathbf{x}^{\ell}) + \mathbf{x}^{\ell}, \mathcal{G}^{\ell}(\mathbf{x}^{\ell})\in\{0,1\}.
  \setlength{\belowdisplayskip}{3pt}
\end{equation}


SkipNet \cite{wang2018skipnet} and convolutional network with adaptive inference graph (Conv-AIG) \cite{veit2018convolutional} are two typical approaches to enabling dynamic layer skipping. Both methods induce lightweight computational overheads to efficiently produce the binary decisions on whether skipping the calculation of a residual block. Specifically, Conv-AIG utilizes two FC layers in each residual block, while the gating function in SkipNet is implemented as an RNN for parameter sharing.

\begin{figure*}
  \centering
  \vspace{-2ex}
    \includegraphics[width=0.85\linewidth]{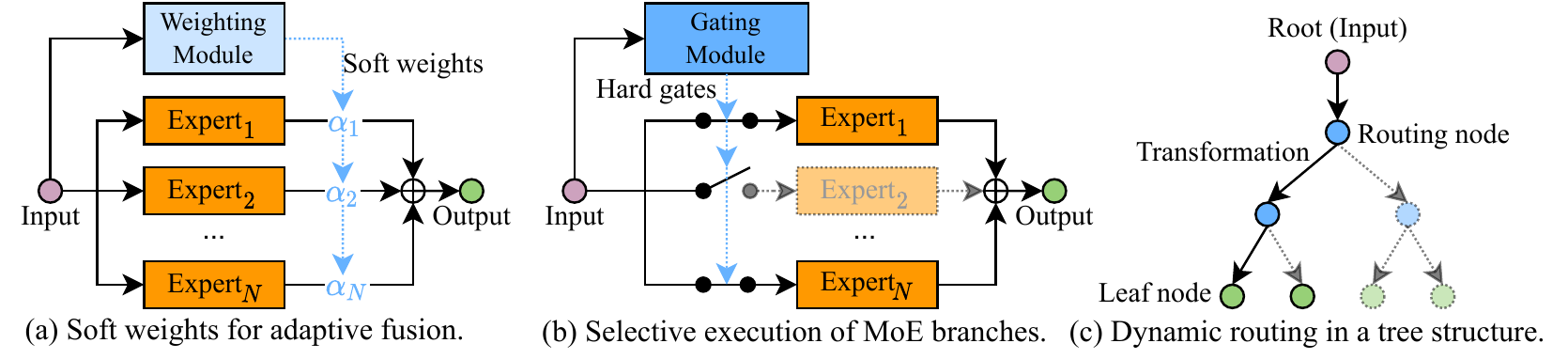}
    \vskip -0.175in
    \caption{MoE with soft weighting (a) and hard gating (b) schemes both adopt an auxiliary module to generate the weights or gates. In the tree structure (c), features (nodes) and transformations (paths) are represented as circles and lines with arrows respectively. Only the full lines are activated.}
    \label{multi_branch}
    \vspace{-3ex}
\end{figure*}

Rather than skipping layers in classic ResNets, dynamic recursive network \cite{Guo_2019_CVPR} iteratively executes one block with shared parameters in each stage. Although the weight-sharing scheme seems similar to the aforementioned IamNN \cite{leroux_iamnn_2018}, the skipping policy of \cite{Guo_2019_CVPR} differs significantly: gating modules are exploited to decide the recursion depth.

Instead of either skipping a layer, or executing it thoroughly with a full numerical precision, a line of work \cite{yu2019any,jin2020adabits} studies adaptive \emph{bit-width} for different layers conditioned on the \emph{resource budget}. Furthermore, fractional skipping \cite{shen2020fractional} adaptively selects a bit-width for each residual block by a gating function based on \emph{input features}.

c) {\emph{Policy network} can be built to take in an input sample, and directly produces the skipping decisions for all the layers in a backbone network \cite{wu2018blockdrop} (see \figurename~\ref{skipping_layers} (c)).}



\vspace{-1.5ex}
\subsubsection{Dynamic Width}\label{dynamic_width}
In addition to dynamic network \emph{depth} (Sec. \ref{dynamic_depth}), a finer-grained form of conditional computation is performing inference with dynamic \emph{width}: although every layer is executed, its multiple units (e.g. neurons, channels or branches) are selectively activated conditioned on the input.

\noindent\textbf{1) {Skipping neurons in fully-connected (FC) layers.}} The computational cost of an FC layer is determined by its input and output dimensions. It is commonly believed that different neuron units are responsible for representing different features, and therefore not all of them need to be activated for every sample. Early studies learn to adaptively control the neuron activations by auxiliary branches \cite{bengio2013estimating,cho2014exponentially,bengio2015conditional} or other techniques such as low-rank approximation \cite{davis2013low}.

\noindent\textbf{2) {Skipping branches in mixture-of-experts (MoE).}} In Sec. \ref{dynamic_depth}, adaptive model ensemble is achieved via a \emph{cascading} way, and later networks are conditionally executed based on early predictions. An alternative approach to improving the model capacity is the MoE \cite{jacobs1991adaptive,eigen2013learning} structure, which means that multiple network branches are built as experts in \emph{parallel}. 
These experts could be selectively executed, and their outputs are fused with data-dependent weights.

Conventional \emph{soft} MoEs \cite{jacobs1991adaptive,eigen2013learning,ma2018modeling} adopt real-valued weights to dynamically rescale the representations obtained from different experts (\figurename~\ref{multi_branch} (a)). In this way, all the branches still need to be executed, and thus the computation cannot be reduced at test time. \emph{Hard} gates with only a fraction of non-zero elements are developed to increase the inference efficiency of the MoE structure (see \figurename~\ref{multi_branch} (b)) \cite{teja2018hydranets,pmlr-v115-wang20d,caidynamic}: let $\mathcal{G}$ denote a gating module whose output is a $N$-dimensional vector $\bm{\alpha}$ controlling the execution of $N$ experts $\mathcal{F}_1,\mathcal{F}_2, \cdots, \mathcal{F}_N$, the final output can be written as
\begin{equation} \label{eq_moe}
  \setlength{\abovedisplayskip}{3pt}
  \mathbf{y} = \sum\nolimits_{n=1}^N [\mathcal{G}(\mathbf{x})]_n \mathcal{F}_n (\mathbf{x}) = \sum\nolimits_{n=1}^N \alpha_n \mathcal{F}_n (\mathbf{x}),
  \setlength{\belowdisplayskip}{3pt}
\end{equation}
and the $n$-th expert will not be executed if $\alpha_n\! =\!0$. 

Hard MoE has been implemented in diverse network structures. For example, HydraNet \cite{teja2018hydranets} replaces the convolutional blocks in a CNN by multiple branches, and selectively execute them conditioned on the input. For another example, dynamic routing network (DRNet) \cite{caidynamic} performs a branch selection in each cell structure which is commonly used in NAS \cite{liu_darts_2018}. On NLP tasks, sparely gated MoE \cite{shazeer2017outrageously} and switch Transformer \cite{2021arXiv210103961F} embeds hard MoE in a long short-term memory (LSTM) \cite{hochreiter1997long} network and a Transformer \cite{vaswani2017attention}, respectively. Instead of making choice with \emph{binary} gates as in \cite{caidynamic}, only the branches corresponding to the \emph{top-K} elements of the real-valued gates are activated in \cite{teja2018hydranets,shazeer2017outrageously,2021arXiv210103961F}.


\noindent\textbf{{3) Skipping channels in CNNs.}}
Modern CNNs usually have considerable channel redundancy. Based on the common belief that the same feature channel can be of disparate importance for different samples, adaptive width of CNNs could be realized by dynamically activating convolutional channels. Compared to the \emph{static} pruning methods \cite{huang2018condensenet,liu2017learning} which remove "unimportant" filters permanently, such a \emph{data-dependent} pruning approach improves the inference efficiency without degrading the model capacity.

a) \emph{Multi-stage architectures along the channel dimension.} Recall that the early-exiting networks \cite{huang2017multi,yang_resolution_2020} discussed in Sec. \ref{dynamic_depth} can be viewed as multi-stage models along the \emph{depth} dimension, where late stages are conditionally executed based on early predictions. One can also build multi-stage architectures along the \emph{width} (channel) dimension, and progressively execute these stages on demand.

Along this direction, an optimal architecture is searched among multiple structures with different widths, and any sample can be output at an early stage when a confident prediction is obtained \cite{yuan2019s2dnas}. Channel gating network (CGNet) \cite{hua2019channel} first executes a subset of convolutional filters in every layer, and the remaining filters are only activated on strategically selected areas.

b) \emph{Dynamic pruning with gating functions.} In the aforementioned progressive activation paradigm, the execution of a later stage is decided based on previous output. As a result, a complete forward propagation is required for every stage, which might be suboptimal for reducing the practical inference latency. Another prevalent solution is to decide the execution of channels at every layer by gating functions. For example, runtime neural pruning (RNP) \cite{lin2017runtime} models the layer-wise pruning as a Markov decision process, and an RNN is used to select specific channel groups at every layer. Moreover, pooling operations followed by FC layers are utilized to generate {\emph{channel-wise hard attention} (i.e. making discrete decisions on whether to activate each channel)} for each sample \cite{gao2018dynamic, herrmann2018end,bejnordi2019batch,chen2019self}. The recent work \cite{li2021dynamic_slimmable} uses a gate module to decide the width for a whole stage of a ResNet. Different reparameterization and optimizing techniques are required for training these gating functions, which will be reviewed in Sec. \ref{training}.


Rather than placing plug-in gating modules \emph{inside} a CNN, GaterNet \cite{chen2019you} builds an \emph{extra} network, which takes in the input sample and directly generates all the channel selection decisions for the backbone CNN. This implementation is similar to BlockDrop \cite{wu2018blockdrop} that exploits an additional policy network for dynamic layer skipping (Sec. \ref{dynamic_depth}).

c) \emph{Dynamic pruning based on feature activations directly} has also been realized without auxiliary branches and computational overheads \cite{liu2019learning}, where a regularization item is induced in training to encourage the sparsity of features.

{On basis of the existing literature on dynamically skipping either network \emph{layers} \cite{wang2018skipnet,veit2018convolutional} or convolutional \emph{filters} \cite{lin2017runtime,gao2018dynamic, herrmann2018end,bejnordi2019batch}, recent work \cite{wang_dual_2020,xia2020fully,bejnordi2020dynamic} has realized dynamic inference with respect to network \emph{depth} and \emph{width} simultaneously: only if a layer is determined to be executed, its channels will be selectively activated, leading to a more flexible adaptation of computational graphs.}

\vspace{-1ex}
\subsubsection{Dynamic Routing} \label{SuperNets}
The aforementioned methods mostly adjust the depth (Sec. \ref{dynamic_depth}) or width (Sec. \ref{dynamic_width}) of \emph{classic architectures} by activating their computational units (e.g. layers \cite{wang2018skipnet,veit2018convolutional} or channels \cite{lin2017runtime,bejnordi2019batch}) conditioned on the input. {Another line of work develops different forms of SuperNets with various possible inference paths, and performs dynamic routing inside the SuperNets to adapt the computational graph to each sample.}

To achieve dynamic routing, there are typically routing nodes that are responsible for allocating features to different paths. For node $s$ at layer $\ell$, let $\alpha_{s\rightarrow j}^{\ell}$ denote the probability of assigning the reached feature $\mathbf{x}^{\ell}_s$ to node $j$ at layer $\ell+1$, the path $\mathcal{F}_{s\rightarrow j}^{\ell}$ will be activated only when $\alpha_{s\rightarrow j}^{\ell}\!>\!0$. The resulting feature reaching node $j$ is represented by
\begin{equation}\label{eq_supernet}
  \setlength{\abovedisplayskip}{3pt}
  \mathbf{x}^{\ell+1}_j = \sum\nolimits_{s\in\left\{s: \alpha_{s\rightarrow j}^{\ell}>0\right\}} \alpha_{s\rightarrow j}^{\ell}\mathcal{F}_{s\rightarrow j}^{\ell}(\mathbf{x}^{\ell}_s).
  \setlength{\belowdisplayskip}{3pt}
\end{equation}

The probability $\alpha_{s\rightarrow j}^{\ell}$ can be obtained in different manners. 
Note that the dynamic early-exiting networks \cite{huang2017multi,yang_resolution_2020} are a special form of SuperNets, where the routing decisions are only made at intermediate classifiers. The CapsuleNets \cite{sabour2017dynamic, e2018matrix} also perform dynamic routing between capsules, i.e. groups of neurons, to character the relations between (parts of) objects. Here we mainly focus on specific architecture designs of the SuperNets and their routing policies.

\noindent\textbf{1) Path selection in multi-branch structures.} The simplest dynamic routing can be implemented by selectively executing \emph{one} of multiple candidate modules at each layer \cite{odena2017changing,liu2018dynamic}, which is equivalent to producing a one-hot probability distribution $\alpha_{s\rightarrow \cdot}^{\ell}$ in Eq. \ref{eq_supernet}. The main difference of this approach to hard MoE (\figurename~\ref{multi_branch} (b)) is that only one branch is activated without any fusion operations.

\noindent\textbf{2) Neural trees and tree-structured networks.} 
As decision trees always perform inference along one forward path that is dependent on input properties, combining tree structure with neural networks can naturally enjoy the adaptive inference paradigm and the representation power of DNNs simultaneously. Note that in a tree structure, the outputs of different nodes are routed to \emph{independent} paths rather than being \emph{fused} as in MoE structures (compare \figurename~\ref{multi_branch} (b), (c)).


a) \emph{Soft decision tree} (SDT) \cite{rota2014neural,kontschieder_deep_2015,frosst2017distilling} adopts neural units as routing nodes (blue nodes in \figurename~\ref{multi_branch} (c)), which decides the portion that the inputs are assigned to their left/right sub-tree. Each leaf node generates a probability distribution over the output space, and the final prediction is the expectation of the results from all leaf nodes. Although the probability for a sample reaching each leaf node in an SDT is data-dependent, all the paths are still executed, which limits the inference efficiency.

b) \emph{Neural trees with deterministic routing policies} \cite{hehn2019end,hazimeh_tree_2020} are designed to make hard routing decisions during inference, avoiding computation on those unselected paths.

c) \emph{Tree-structured DNNs.} Instead of developing decision trees containing neural units, a line of work builds special network architectures to endow them with the routing behavior of decision trees. For instance, a small CNN is first executed to classify each sample into coarse categories, and specific sub-networks are conditionally activated based on the coarse predictions \cite{yan2015hd}. A subsequent work \cite{ioannou_decision_2016} not only partitions samples to different sub-networks, but also divides and routes the feature channels.

Different to those networks using neural units only in routing nodes \cite{hehn2019end,hazimeh_tree_2020}, or routing each sample to pre-designed sub-networks \cite{yan2015hd,ioannou_decision_2016}, adaptive neural tree (ANT) \cite{tanno_adaptive_2019} adopts CNN modules as feature transformers in a hard neural tree (see lines with arrows in \figurename~\ref{multi_branch} (c)), and learns the tree structure together with the network parameters simultaneously in the training stage.

\noindent\textbf{3) Others.} 
{Performing dynamic routing within more general SuperNet architectures is also a recent research trend. Representatively, an architecture distribution with partly shared parameters is \emph{searched} from a SuperNet containing $\sim \!\! 10^{25}$ sub-networks \cite{cheng2020instanas}. During inference, every sample is allocated by a controller network to one sub-network with appropriate computational cost. Instead of training a standalone controller network, gating modules are plugged inside the \emph{hand-designed} SuperNet (see \figurename~\ref{multi_scale} (d)) to decide the routing path based on intermediate features \cite{li_learning_2020}. 



\begin{figure*}
  \centering
  \vspace{-2ex}
    \includegraphics[width=0.9\linewidth]{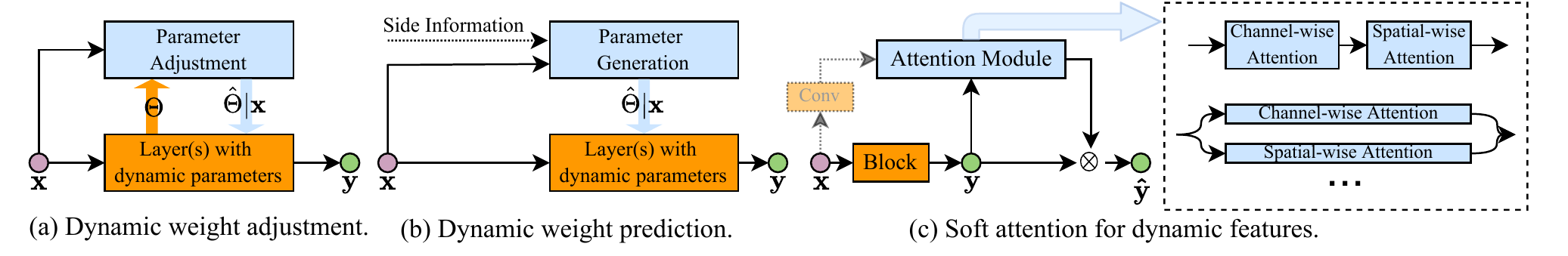}
    \vskip -0.2in
    \caption{{Three implementations of dynamic parameters: adjusting (a) or generating (b) the backbone parameters based on the input, and (c) dynamically rescaling the features with the attention mechanism.}}
    \label{dynamic_params}
    \vskip -0.2in
\end{figure*}

\vspace{-1.5ex}
\subsection{Dynamic Parameters} \label{adaptive_params}
\vspace{-0.25ex}
Although the networks with dynamic \emph{architectures} in Sec. \ref{dynamic_arch} can adapt their inference graphs to each sample and achieve an efficient allocation of computation, they usually have special architecture designs, requiring specific training strategies or careful hyper-parameters tuning (Sec. \ref{sec:discussion}).

Another line of work adapts network \emph{parameters} to different inputs while keeping the architectures fixed, which has been shown effective in improving the representation power of networks with a minor increase of computational cost. Given an input sample $\mathbf{x}$, 
the output of a conventional network (module) with static parameters can be written as $\mathbf{y}\! =\!\mathcal{F}(\mathbf{x},\bm{\Theta})$. In contrast, the output of a model with dynamic parameters could be represented by
\begin{equation}
  \setlength{\abovedisplayskip}{3pt}
  \mathbf{y} = \mathcal{F}(\mathbf{x},\bm{\hat{\Theta}}|\mathbf{x}) = \mathcal{F}(\mathbf{x},\mathcal{W}(\mathbf{x}, \bm{\Theta})),
  \label{eq:dynamic_parameter}
  \setlength{\belowdisplayskip}{3pt}
\end{equation}
where $\mathcal{W}(\cdot, \bm{\Theta})$ is the operation producing input-dependent parameters, and its design has been extensively explored.

In general, the parameter adaptation can be achieved from three aspects (see \figurename~\ref{dynamic_params}): 1) adjusting the trained parameters based on the input (Sec. \ref{dynamic_param_adjust}); 2) directly generating the network parameters from the input (Sec. \ref{weight_predict}); and 3) rescaling the features with soft attention (Sec. \ref{sec:attention}).

\vspace{-1ex}
\subsubsection{Parameter Adjustment}
\label{dynamic_param_adjust}
\vspace{-0.5ex}
A typical approach to parameter adaptation is adjusting the weights based on their input during inference as presented in \figurename~\ref{dynamic_params} (a). This implementation usually evokes little computation to obtain the adjustments, e.g., attention weights \cite{harley_segmentation-aware_2017, su_pixel-adaptive_2019,yang2019condconv,chen_dynamic_2020_attentionOver} or sampling offsets \cite{dai2017deformable,zhu_deformable_2019,gao_deformable_2019}.

\noindent\textbf{1) Attention on weights.} {To improve the representation power without noticeably increasing the computation, 
soft attention can be performed on multiple convolutional kernels, producing an adaptive ensemble of parameters \cite{yang2019condconv,chen_dynamic_2020_attentionOver}.}
Assuming that there are $N$ kernels $\mathbf{W}_n, n\! =1,2,\!\cdots\!,N$, such a dynamic convolution 
can be formulated as 
\begin{equation}
  \setlength{\abovedisplayskip}{3pt}
\mathbf{y} = \mathbf{x} \star \mathbf{\tilde{W}} = \mathbf{x} \star (\sum\nolimits_{n=1}^N \alpha_n\mathbf{W}_n). 
\setlength{\belowdisplayskip}{3pt}
\end{equation}
This procedure increases the model capacity yet remains high efficiency, as the result obtained through fusing the outputs of $N$ convolutional branches (as in MoE structures, see \figurename~\ref{multi_branch} (a)) is equivalent to that produced by performing once convolution with $\mathbf{\tilde{W}}$. However, only $\sim\!1/N$ times of computation is consumed in the latter approach.
 


Weight adjustment could also be achieved by performing soft attention over the \emph{spatial locations} of convolutional weights \cite{harley_segmentation-aware_2017, su_pixel-adaptive_2019}. For example, segmentation-aware convolutional network \cite{harley_segmentation-aware_2017} applies locally masked convolution to aggregate information with larger weights from similar pixels, which are more likely to belong to the same object.
Unlike \cite{harley_segmentation-aware_2017} that requires a sub-network for feature embedding, pixel-adaptive convolution (PAC) \cite{su_pixel-adaptive_2019} adapts the convolutional weights based on the attention mask generated from the input feature at each layer.

{Instead of adjusting weights conditioned on every sample itself, meta-neighborhoods \cite{shan2020meta} adapt the network parameters to each input sample based on its similarity to the neighbors stored in a dictionary.}

\noindent\textbf{2) Kernel shape adaptation.} 
Apart from adaptively scaling the weight \emph{values}, parameter adjustment can also be realized to reshape the convolutional kernels and achieve \emph{dynamic reception of fields}. Towards this direction, 
deformable convolutions \cite{dai2017deformable,zhu_deformable_2019} sample feature pixels from adaptive locations when performing convolution on each pixel. Deformable kernels \cite{gao_deformable_2019} samples weights in the kernel space to adapt the \emph{effective} reception field (ERF) while leaving the reception field unchanged. Table \ref{tab:deform_kernels} summarizes the formulations of the above three methods. 
{Due to their irregular memory access and computation pattern, these kernel shape adaptation approaches typically require customized CUDA kernels for the implementation on GPUs. However, recent literature has shown that the practical efficiency of deformable convolution could be effectively improved by co-designing algorithm and hardware based on embedded devices such as FPGAs \cite{huang2021codenet}.}
\begin{table*}
  \vspace{-2ex}
  \caption{{Kernel shape adaptation by dynamically sampling feature pixels \cite{dai2017deformable,zhu_deformable_2019} or convolutional weights \cite{gao_deformable_2019}.}}
  \vspace{-4ex}
  \label{tab:deform_kernels}
  \begin{center}
    \begin{tabular}{c|c|c|c}
      \hline
      \textbf{Method} & \textbf{Formulation} & \textbf{Sampled Target} & \textbf{Dynamic Mask} \\
      \hline
      Regular Convolution & $\mathbf{y(p)} = \sum\nolimits_{k=1}^K \mathbf{W}(\mathbf{p}_k) \mathbf{x}(\mathbf{p+p}_k)$ & - & -  \\
      \hline
      Deformable ConvNet-v1 \cite{dai2017deformable} & $\mathbf{y(p)} = \sum\nolimits_{k=1}^K \mathbf{W}(\mathbf{p}_k) \mathbf{x}(\mathbf{p+p}_k+\Delta \mathbf{p}_k)$ & Feature map & No  \\
      Deformable ConvNet-v2 \cite{zhu_deformable_2019} & $\mathbf{y(p)} = \sum\nolimits_{k=1}^K \mathbf{W}(\mathbf{p}_k) \mathbf{x}(\mathbf{p+p}_k+\Delta \mathbf{p}_k)\Delta \mathbf{m}_k$ & Feature map & Yes  \\
      Deformable Kernels \cite{gao_deformable_2019} & $\mathbf{y(p)} = \sum\nolimits_{k=1}^K \mathbf{W}(\mathbf{p}_k+\Delta \mathbf{p}_k) \mathbf{x}(\mathbf{p+p}_k)$ & Conv kernel & No \\
      \hline
    \end{tabular}
  \end{center}
  \vspace{-5ex}
\end{table*}

\vspace{-1ex}
\subsubsection{Weight Prediction} \label{weight_predict}
\vspace{-0.25ex}
Compared to making modifications on model parameters on the fly (Sec. \ref{dynamic_param_adjust}), weight prediction \cite{denil_predicting_2013} is more straightforward: it directly generates (a subset of) input-adaptive parameters with an independent model at test time (see \figurename~\ref{dynamic_params} (b)). This idea was first suggested in \cite{schmidhuber1992learning}, where both the weight prediction model and the backbone model were feedforward networks. Recent work has further extended the paradigm to modern network architectures and tasks.

\noindent\textbf{1) General architectures.} Dynamic filter networks (DFN) \cite{jia2016dynamic} and HyperNetworks \cite{ha2016hypernetworks} are two classic approaches realizing runtime weight prediction for CNNs and RNNs, respectively. Specifically, a filter generation network is built in DFN \cite{jia2016dynamic} to produce the filters for a convolutional layer. As for processing sequential data (e.g. a sentence), the weight matrices of the main RNN are predicted by a smaller one at each time step conditioned on the input (e.g. a word) \cite{ha2016hypernetworks}. WeightNet \cite{ma_weightnet_2020} unifies the dynamic schemes of \cite{yang2019condconv} and \cite{hu2018squeeze} by predicting the convolutional weights via simple grouped FC layers, achieving competitive results in terms of the accuracy-FLOPs\footnote{{Floating point operations, which is widely used as a measure of inference efficiency of deep networks.}} and accuracy-parameters trade-offs.

{Rather than generating standard \emph{convolutional} weights, LambdaNetworks \cite{bello2021lambdanetworks} learns to predict the weights of \emph{linear} projections based on the contexts of each pixel together with the relative position embeddings, showing advantages in terms of computational cost and memory footprint.}


\noindent\textbf{2) Task-specific information} 
has also been exploited to predict model parameters on the fly, enabling dynamic networks to generate task-aware feature embeddings. For example, edge attributes are utilized in \cite{simonovsky_dynamic_2017} to generate filters for graph convolution, and camera perspective is incorporated in \cite{kang2017incorporating} to generate weights for image convolution. {Such task-aware weight prediction has been shown effective in improving the data efficiency on many tasks, including visual question answering \cite{de2017modulating,perez2018film} and few-shot learning \cite{bertinetto2016learning,wang2019tafe}.}

\vspace{-1ex}
\subsubsection{Dynamic Features} \label{sec:attention}
\vspace{-0.25ex}
The main goal of either \emph{adjusting} (Sec. \ref{dynamic_param_adjust}) or \emph{predicting} (Sec. \ref{weight_predict}) model parameters is producing more dynamic and informative features, and therefore enhancing the representation power of deep networks. A more straightforward solution is rescaling the features with input-dependent soft attention (see \figurename~\ref{dynamic_params} (c)), which requires minor modifications on computational graphs. Note that for a linear transformation $\mathcal{F}$, applying attention $\bm{\alpha}$ on the output is equivalent to performing computation with re-weighted parameters, i.e.
\begin{equation}\label{dynamic_feature_key}
  \setlength{\abovedisplayskip}{3pt}
    \mathcal{F}(\mathbf{x},\bm{\Theta})\otimes \bm{\alpha}= \mathcal{F}(\mathbf{x},\bm{\Theta}\otimes\bm{\alpha}).
    \setlength{\belowdisplayskip}{3pt}
\end{equation}


\noindent\textbf{1) Channel-wise attention} is one of the most common soft attention mechanisms. Existing work typically follows 
the form in squeeze-and-excitation network (SENet) \cite{hu2018squeeze}:
\begin{equation}\label{se_attention}
  \setlength{\abovedisplayskip}{3pt}
\mathbf{\tilde{y}}\! = \!\mathbf{y} \otimes \bm{\alpha}\! =\!\mathbf{y} \otimes \mathcal{A}(\mathbf{y}),  \bm{\alpha}\in\left[0,1\right]^C.
\setlength{\belowdisplayskip}{3pt}
\end{equation} 
In Eq. \ref{se_attention}, $\mathbf{y}\! =\!\mathbf{x} \star \mathbf{W}$ is the output feature of a convolutional layer with $C$ channels, and $\mathcal{A}(\cdot)$ is a lightweight function composed of pooling and linear layers for producing $\bm{\alpha}$.
Taking the convolution into account, the procedure can also be written as $\mathbf{\tilde{y}}\! = \!(\mathbf{x} \star \mathbf{W}) \otimes \bm{\alpha}\! =\! \mathbf{x} \star (\mathbf{W} \otimes \bm{\alpha})$, from which we can observe that applying attention on features is equivalent to performing convolution with dynamic weights.


Other implementations for attention modules have also been developed, including using standard deviation to provide more statistics \cite{lee2019srm}, or replacing FC layers with efficient 1D convolutions \cite{wang_eca-net_2020}.
The empirical performance of three computational graphs for soft attention is studied in \cite{guo_spanet_2020}: 1) $\mathbf{\tilde{y}}\! =\!\mathbf{y}\otimes \mathcal{A}(\mathbf{y})$, 2) $\mathbf{\tilde{y}}\! =\!\mathbf{y}\otimes \mathcal{A}(\mathbf{x})$ and 3) $\mathbf{\tilde{y}}\! =\!\mathbf{y}\otimes \mathcal{A}(\mathrm{Conv}(\mathbf{x}))$. It is found that the three forms yield different performance in different backbone networks.

\noindent\textbf{2) Spatial-wise attention}. Spatial locations in features could also be dynamically rescaled with attention to improve the representation power of deep models \cite{wang2017residual}. Instead of using pooling operations to efficiently gather global information as in channel-wise attention, convolutions are often adopted in spatial-wise attention to encode local information. Moreover, these two types of attention modules can be integrated in one framework \cite{roy2018concurrent,chen_sca-cnn_2017, woo_cbam_2018,hu2018gather} (see \figurename~\ref{dynamic_params} (c)).

\noindent\textbf{3) Dynamic activation functions.} The aforementioned approaches to generating dynamic features usually apply soft attention before static activation functions. A recent line of work has sought to increase the representation power of models with dynamic activation functions \cite{chen_dynamic_2020_relu,ma2020funnel}. For instance, 
DY-ReLU \cite{chen_dynamic_2020_relu} replaces ReLU $(\mathbf{y}_c\! =\!\max (\mathbf{x}_c, 0))$ with the max value among $N$ linear transformations $\mathbf{y}_c\! =\!\max_{n} \left\{a_c^n \mathbf{x}_c + b_c^n\right\}$, where $c$ is the channel index, and $a_c^n, b_c^n$ are linear coefficients calculated from $\mathbf{x}$. On many vision tasks, these dynamic activation functions can effectively improve the performance of different network architectures with negligible computational overhead.

To summarize, soft attention has been exploited in many fields due to its simplicity and effectiveness. Moreover, it can be incorporated with other methods conveniently. E.g., by replacing the weighting scalar $\alpha_n$ in Eq. \ref{eq_moe} with channel-wise \cite{li_selective_2019} or spatial-wise \cite{wang_autoscaler_2016} attention, the output of multiple branches with independent kernel sizes \cite{li_selective_2019} or feature resolutions \cite{wang_autoscaler_2016} are adaptively fused.

Note that we leave out the detailed discussion on the self attention mechanism, which is widely studied in both NLP \cite{vaswani2017attention, devlin_bert_2019} and CV fields \cite{wang2018non,yue2018compact,dosovitskiy2020image} to re-weight features based on the similarity between queries and keys at different locations (temporal or spatial). 
Readers who are interested in this topic may refer to review studies \cite{chaudhari2019attentive,zhu2019empirical,khan2021transformers}. In this survey, we mainly focus on the feature re-weighting scheme in the framework of dynamic inference. 

\vspace{-2ex}
\section{Spatial-wise Dynamic Networks}
\vspace{-0.25ex}
\label{sec_spatially_adaptive}
In visual learning, it has been found that not all locations contribute equally to the final prediction of CNNs \cite{zhou2016learning}, which suggests that \emph{spatially} dynamic computation has great potential for reducing computational redundancy.
In other words, making a correct prediction may only require processing a fraction of pixels or regions with an adaptive amount of computation. Moreover, based on the observations that low-resolution representations are sufficient to yield decent performance for most inputs \cite{howard2017mobilenets}, the static CNNs that take in all the input with the same resolution may also induce considerable redundancy.

To this end, spatial-wise dynamic networks are built to perform adaptive inference with respect to different spatial locations of images.
According to the granularity of dynamic computation, we further categorize the relevant approaches into three levels: {\emph{pixel level} (Sec. \ref{pixel_level}), \emph{region level} (Sec. \ref{region_level}) and \emph{resolution level} (Sec. \ref{dynamic_resolution}).}



\vspace{-2ex}
\subsection{Pixel-level Dynamic Networks}
\vspace{-0.25ex}
\label{pixel_level}
Commonly seen spatial-wise dynamic networks perform adaptive computation at the pixel level. Similar to the categorization in Sec. \ref{sec_sample_wise}, pixel-level dynamic networks are grouped into two types: models with pixel-specific \emph{dynamic architectures}  (Sec. \ref{pixel_dynamic_arch}) and \emph{dynamic parameters} (Sec. \ref{pixel_dynamic_param}).

\vspace{-1ex}
\subsubsection{Pixel-wise Dynamic Architectures} \label{pixel_dynamic_arch}
\vspace{-0.25ex}
Based on the common belief that foreground pixels are more informative and computational demanding than those in the background, some dynamic networks learn to adjust their architectures for each pixel. Existing literature generally achieves this by 1) \emph{dynamic sparse convolution}, which only performs convolutions on a subset of sampled pixels; 2) \emph{additional refinement}, which strategically allocates extra computation (e.g. layers or channels) on certain spatial positions.

\begin{figure}
  \centering
  \vspace{-1ex}
    \includegraphics[width=\linewidth]{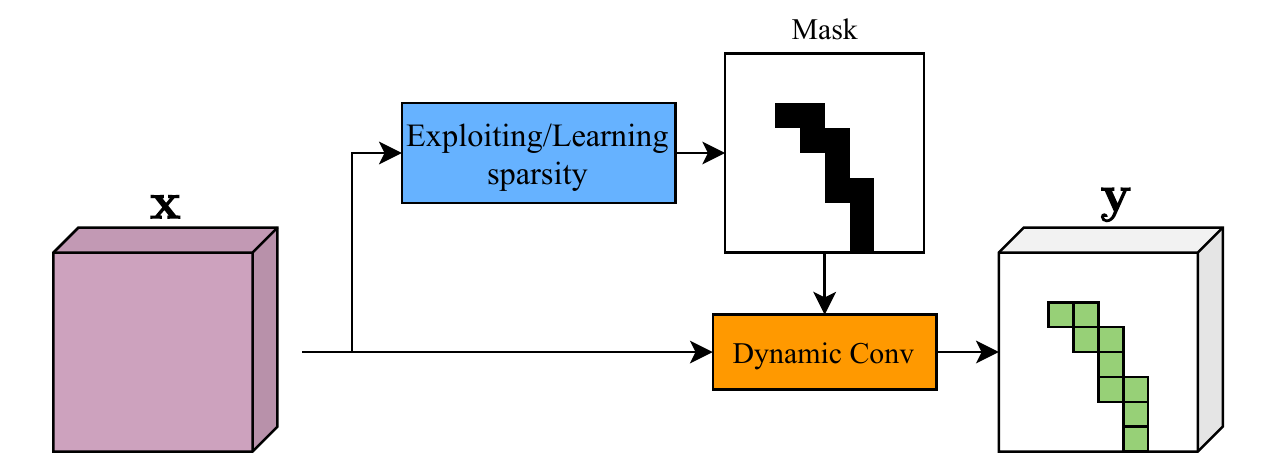}
    \vskip -0.15in
    \caption{{Dynamic convolution on selected spatial locations. The 1 elements (black) in the spatial mask determine the pixels (green) that require computation in the output feature map.}}
    \label{fig_spatial_location_specific}
    \vspace{-3ex}
\end{figure}
\noindent\textbf{1) Dynamic sparse convolution.} To reduce the unnecessary computation on less informative locations, convolution can be performed only on strategically sampled pixels. Existing sampling strategies include 1) making use of the intrinsic sparsity of the input \cite{ren_sbnet_2018}; 2) predicting the positions of zero elements on the output \cite{dong_more_2017,cao_seernet_2019}; and 3) estimating the saliency of pixels \cite{kong_pixel-wise_2019,verelst_dynamic_2020,xie_spatially_2020}. A typical approach is using an extra branch to generate a spatial mask, determining the execution of convolution on each pixel (see \figurename~\ref{fig_spatial_location_specific}). {Pixel-wise dynamic depth could also be achieved based on a halting scheme \cite{figurnov2017spatially}} (see Sec. \ref{dynamic_depth}). These dynamic convolutions usually neglect the unselected positions, which might degrade the network performance. Interpolation is utilized in \cite{xie_spatially_2020} to efficiently fill those locations, therefore alleviating the aforementioned disadvantage.


\noindent\textbf{2) Dynamic additional refinement.} Instead of only sampling certain pixels to perform convolutions, another line of work first conducts relatively cheap computation on the whole feature map, and adaptively activate extra modules on selected pixels for further \emph{refinement}. Representatively, dynamic capacity network \cite{almahairi2016dynamic} generates coarse features with a shallow model, and salient pixels are sampled based on the gradient information. 
For these salient pixels, extra layers are applied to extract finer features. Similarly, specific positions are additionally processed by a fraction of convolutional filters in \cite{hua2019channel}. These methods adapt their network architectures in terms of \emph{depth} or \emph{width} at the pixel level, achieving a spatially adaptive allocation of computation.

{The aforementioned dynamic additional refinement approaches \cite{almahairi2016dynamic,hua2019channel} are mainly developed for image classification.} On the semantic segmentation task, pixel-wise \emph{early exiting} (see also Sec. \ref{dynamic_depth}) is proposed in \cite{li_not_2017}, where the pixels with high prediction confidence are output without being processed by deeper layers. PointRend \cite{kirillov2020pointrend} shares a similar idea, and applies additional FC layers on selected pixels with low prediction confidence, which are more likely to be on borders of objects. All these researches demonstrate that by exploiting the spatial redundancy in image data, dynamic computation at the pixel level beyond sample level significantly increases the model efficiency.

\vspace{-1ex}
\subsubsection{Pixel-wise Dynamic Parameters} \label{pixel_dynamic_param}
\vspace{-0.25ex}
In contrast to entirely skipping the convolution operation on a subset of pixels, dynamic networks can also apply data-dependent parameters on different pixels for improved representation power or adaptive reception fields.

\noindent\textbf{1) Dynamic weights.} {Similar to the sample-wise dynamic parameter methods (Sec. \ref{adaptive_params}), pixel-level dynamic weights are achieved by test-time \emph{adjustment} \cite{harley_segmentation-aware_2017, su_pixel-adaptive_2019}, \emph{prediction} \cite{bhowmik_training-free_2017,wu_dynamic_2018,hu_meta-sr_2019,wang_carafe_2019} or \emph{dynamic features} \cite{wang_autoscaler_2016,roy2018concurrent,chen_sca-cnn_2017,woo_cbam_2018}. Take weight prediction as an example,} typical approaches generate an $H\!\times\!W\!\times\! k^2$ kernel map to produce spatially dynamic weights ($H,W$ are the spatial size of the output feature and $k$ is the kernel size). Considering the pixels belonging to the same object may share identical weights, dynamic region-aware convolution (DRConv) \cite{chen_dynamic_2020} generates a segmentation mask for an input image, dividing it into $m$ regions, for each of which a weight generation network is responsible for producing a data-dependent kernel.

\noindent\textbf{2) Dynamic reception fields.} Traditional convolution operations usually have a fixed shape and size of kernels (e.g. the commonly used $3\!\times\!3$ 2D convolution). The resulting uniform reception field across all the layers may have limitations for recognizing objects with varying shapes and sizes. {To tackle this, a line of work learns to adapt the reception field for different feature pixels \cite{dai2017deformable, zhu_deformable_2019, gao_deformable_2019}, as discussed in Sec. \ref{dynamic_param_adjust}.} Instead of adapting the sampling location of features or kernels, adaptive connected network \cite{wang_adaptively_2019} realizes a dynamic trade-off among self transformation (e.g. $1\!\times\!1$ convolution), local inference (e.g. $3\!\times\!3$ convolution) and global inference (e.g. FC layer). The three branches of outputs are fused with data-dependent weighted summation. 
Besides images, the local and global information in non-Euclidean data, such as graphs, could also be adaptively aggregated.

\vspace{-1.5ex}
\subsection{Region-level Dynamic Networks} \label{region_level}
\vspace{-0.5ex}
{Pixel-level dynamic networks mentioned in Sec. \ref{pixel_level} often require specific implementations for sparse computation, and consequently may face challenges in terms of achieving real acceleration on hardware \cite{xie_spatially_2020}.} An alternative approach is performing adaptive inference on \emph{regions/patches} of input images. There mainly exists two lines of work along this direction (see \figurename~\ref{fig_spatial_select_regions}): one performs parameterized \emph{transformations} on a region of feature maps for more accurate prediction (Sec. \ref{dynamic_transofrm}), and the other learns patch-level \emph{hard attention}, with the goal of improving the effectiveness and/or efficiency of models (Sec. \ref{hard_attention_pathces}).

\vspace{-1ex}
\subsubsection{Dynamic Transformations} \label{dynamic_transofrm}
\vspace{-0.25ex}
Dynamic transformations (e.g. affine/projective/thin plate spline transformation) can be performed on images to undo certain variations \cite{jaderberg_spatial_2015} for better generalization ability, or to exaggerate the salient regions \cite{recasens_learning_2018} for discriminative feature representation. For example, spatial transformer \cite{jaderberg_spatial_2015} adopts a localization network to generate the transformation parameters, and then applies the parameterized transformation to recover the input from the corresponding variations. 
Moreover, transformations are learned to adaptively zoom-in the salient regions on some tasks where the model performance is sensitive to a small portion of regions. 
\begin{figure}
  \centering
  \vspace{-1ex}
    \includegraphics[width=\linewidth]{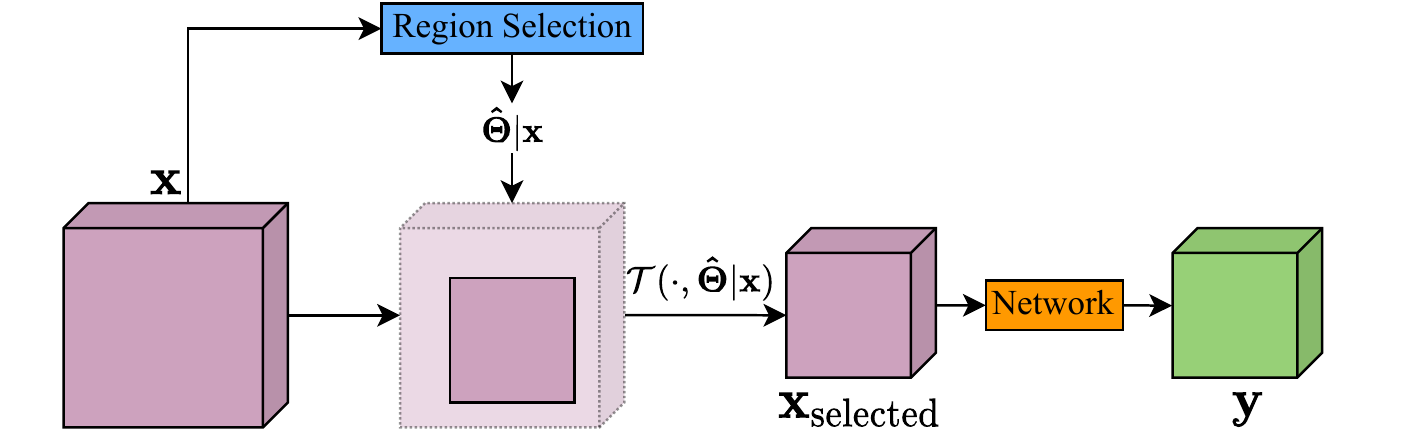}
    \vskip -0.15in
    \caption{{Region-level dynamic inference. The region selection module generates the transformation/localization parameters, and the subsequent network performs inference on the transformed/cropped region.}}
    \label{fig_spatial_select_regions}
    \vspace{-4ex}
\end{figure}
\vspace{-1ex}
\subsubsection{Hard Attention on Selected Patches} \label{hard_attention_pathces}
\vspace{-0.25ex}

Inspired by the fact that informative features may only be contained in certain regions of an image, dynamic networks with hard spatial attention are explored to strategically select patches from the input for improved efficiency.


\noindent\textbf{1) Hard attention with RNNs.} The most typical approach is formulating a classification task as a sequential decision process, {and adopting RNNs to make iterative predictions based on selected patches \cite{mnih_recurrent_2014,li_dynamic_2017}.} For example, images are classified within a fixed number of steps, and at each step, the classifier RNN only sees a cropped patch, deciding the next attentional location until the last step is reached \cite{mnih_recurrent_2014}. An adaptive step number is further achieved by including early stopping in the action space \cite{li_dynamic_2017}. Glance-and-focus network (GFNet) \cite{wang2020glance} builds a general framework of region-level adaptive inference by sequentially focusing on a series of selected patches, and is compatible with most existing CNN architectures. The recurrent attention mechanism together with the early exiting paradigm enables both \emph{spatially} and \emph{temporally} adaptive inference \cite{li_dynamic_2017,wang2020glance}.


\noindent\textbf{2) Hard attention with other implementations.} Rather than using an RNN to predict the region position that the model should pay attention to, class activation mapping (CAM) \cite{zhou2016learning} is leveraged in \cite{rosenfeld_visual_2016} to iteratively focus on salient patches. At each iteration, the selection is performed on the previously cropped input, leading to a progressive refinement procedure. A multi-scale CNN is built in \cite{fu_look_2017}, where the sub-network in each scale takes in the cropped patch from the previous scale, and is responsible for simultaneously producing 1) the feature representations for classification and 2) the attention map for the next scale. {Without an iterative manner, the recent differentiable patch selection \cite{cordonnier2021differentiable} adopts a differentiable top-K module to select a fixed number of patches in one step.}

\vspace{-2ex}
\subsection{Resolution-level Dynamic Networks} \label{dynamic_resolution}
\vspace{-0.5ex}
The researches discussed above typically divide feature maps into different areas (pixel-level or region-level) for adaptive inference.
On a coarser granularity, some dynamic networks could treat each image as a whole by processing feature representations with adaptive resolutions. Although it has been observed that a low resolution might be sufficient for recognizing most "easy" samples \cite{howard2017mobilenets}, conventional CNNs mostly process all the inputs with the same resolution, inducing considerable redundancy. Therefore, resolution-level dynamic networks exploit spatial redundancy from the perspective of feature resolution rather than the saliency of different locations. Existing approaches mainly include 1) scaling the inputs with adaptive ratios (Sec. \ref{adaptive_scaling_ratio}); 2) selectively activating the sub-networks with different resolutions in a multi-scale architecture (Sec. \ref{dynamic_res_multiscale}).

\vspace{-1ex}
\subsubsection{Adaptive Scaling Ratios} \label{adaptive_scaling_ratio}
\vspace{-0.25ex}
Dynamic resolution can be achieved by scaling features with adaptive ratios. For example, a small sub-network is first executed to predict a scale distribution of faces on the face detection task, then the input images are adaptively zoomed, so that all the faces fall in a suitable range for recognition \cite{hao_scale-aware_2017}. A plug-in module is used by \cite{yang_dynamic-stride-net_2019} to predict the stride for the first convolution block in each ResNet stage, producing features with dynamic resolution.

\vspace{-1ex}
\subsubsection{Dynamic Resolution in Multi-scale Architectures} \label{dynamic_res_multiscale} 
\vspace{-0.25ex}
An alternative approach to achieving dynamic resolution is building multiple sub-networks in a parallel \cite{wang_elastic_2019} or cascading \cite{yang_resolution_2020} way. These sub-networks with different feature resolutions are selectively activated conditioned on the input during inference. For instance, Elastic \cite{wang_elastic_2019} realizes a \emph{soft} selection from multiple branches at every layer, where each branch performs a downsample-convolution-upsample procedure with an independent scaling ratio. To practically avoid redundant computation, a \emph{hard} selection is realized by \cite{yang_resolution_2020}, which allows each sample to conditionally activate sub-networks that process feature representations with resolution from low to high (see \figurename~\ref{multi_scale} (c) in Sec. \ref{dynamic_depth}).

\vspace{-2ex}
\begin{figure*}
  \centering
  \vspace{-2ex}
    \includegraphics[width=0.9\linewidth]{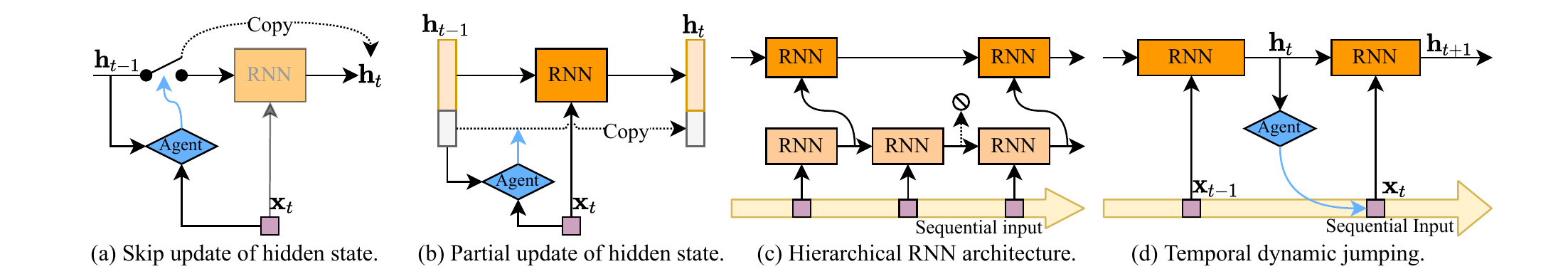}
    \vskip -0.15in
    \caption{{Temporally adaptive inference. The first three approaches dynamically allocate computation in each step by (a) skipping the update, (b) partially updating the state, or (c) conditional computation in a hierarchical structure. The agent in (d) decides where to read in the next step.}}
    \label{fig_temporal_skim}
    \vspace{-3ex}
\end{figure*}
\section{Temporal-wise Dynamic Networks} \label{sec_temporal_adaptive}
\vspace{-0.25ex}
Apart from the spatial dimension (Sec. \ref{sec_spatially_adaptive}), adaptive computation could also be performed along the temporal dimension of sequential data, such as texts (Sec. \ref{temporal_text}) and videos (Sec. \ref{sec:tempoal_video}). {In general, network efficiency can be improved by dynamically allocating less/no computation to the inputs at unimportant temporal locations.}


\vspace{-1.5ex}
\subsection{RNN-based Dynamic Text Processing}
\vspace{-0.25ex}
\label{temporal_text}
Traditional RNNs mostly follow a static inference paradigm, i.e. input tokens are read sequentially to update a hidden state at each time step, which could be written as
\begin{equation}\label{eq_rnn}
  \setlength{\abovedisplayskip}{3pt}
\mathbf{h}_t = \mathcal{F}(\mathbf{x}_t, \mathbf{h}_{t-1}), t=1,2,\cdots, T.
\setlength{\belowdisplayskip}{3pt}
\end{equation}
Such a static inference paradigm induces significant redundant computation, as different tokens usually have different contributions to the downstream tasks.
A type of dynamic RNN is developed for allocating appropriate computational cost at each step. Some learn to \emph{"skim"} unimportant tokens by dynamic update of hidden states (Sec. \ref{skimming_text}), and others conduct \emph{adaptive reading} to avoid processing task-irrelevant tokens. Specifically, such adaptive reading can be achieved by \emph{early exiting} (Sec. \ref{temporal_early_exit}) or \emph{dynamic jumping} (Sec. \ref{text_jumping}).

\vspace{-1ex}
\subsubsection{Dynamic Update of Hidden States}
\vspace{-0.25ex}
\label{skimming_text}
Since not all the tokens are essential for capturing the task-relevant information in a sequence, dynamic RNNs can be built to adaptively update their hidden states at each time step. Less informative tokens will be coarsely \emph{skimmed}, i.e. the states are updated with cheaper computation.

\noindent\textbf{1) Skipping the update.} For unimportant inputs at certain temporal locations, dynamic models can learn to entirely skip the update of hidden states (see \figurename~\ref{fig_temporal_skim} (a)), i.e.
\begin{equation}
  \setlength{\abovedisplayskip}{3pt}
  \mathbf{h}_t = \alpha_t\mathcal{F}(\mathbf{x}_t, \mathbf{h}_{t-1}) + (1-\alpha_t)\mathbf{h}_{t-1}, \alpha_t\in\left\{0,1\right\}.
  \setlength{\belowdisplayskip}{3pt}
\end{equation}
For instance, Skip-RNN \cite{campos_skip_2018} updates a controlling signal in every step to determine whether to update or \emph{copy} the hidden state from the previous step. An extra agent is adopted by Structural-Jump-LSTM \cite{hansen_neural_2019} to make the skipping decision conditioned on the previous state and the current input. Without training the RNNs and the controllers jointly as in \cite{campos_skip_2018} and \cite{hansen_neural_2019}, a predictor is trained in \cite{tao_skipping_2019} to estimate whether each input will make a "significant change" on the hidden state. The update is identified worthy to be executed only when the predicted change is greater than a threshold.

\noindent\textbf{2) Coarse update. } As directly skipping the update may be too aggressive, dynamic models could also update the hidden states with adaptively allocated operations. In specific, a network can adapt its architecture in every step, i.e.  
\begin{equation}
  \setlength{\abovedisplayskip}{3pt}
  \mathbf{h}_t = \mathcal{F}_t(\mathbf{x}_t, \mathbf{h}_{t-1}), t=1,2,\cdots, T,
  \setlength{\belowdisplayskip}{3pt}
\end{equation}
where $\mathcal{F}_t$ is determined based on the input $\mathbf{x}_t$. One implementation is selecting a subset of dimensions of the hidden state to calculate, and copying the remaining from the previous step \cite{jernite_variable_2017,seo_neural_2018}, as shown in \figurename~\ref{fig_temporal_skim} (b). To achieve the partial update, a subset of rows in weight matrices of the RNN is dynamically activated in \cite{jernite_variable_2017}, while Skim-RNN \cite{seo_neural_2018} makes a choice between two independent RNNs.

When the hidden states are generated by a multi-layer network, the update could be interrupted at an intermediate layer based on an accumulated halting score \cite{graves2016adaptive}.

To summarize, a coarse update can be realized by data-dependent network \emph{depth} \cite{graves2016adaptive} or \emph{width} \cite{jernite_variable_2017,seo_neural_2018}. 


{\noindent\textbf{3) Selective updates in hierarchical RNNs.} Considering the intrinsic hierarchical structure of texts (e.g. sentence-word-character), researchers have developed hierarchical RNNs to encode the temporal dependencies with different timescales using a dynamic update mechanism \cite{chung_hierarchical_2017,ke_focused_2018}. During inference, the RNNs at higher levels will selectively update their states conditioned on the output of low-level ones (see \figurename~\ref{fig_temporal_skim} (c)).
For example, when a character-level model in \cite{chung_hierarchical_2017} detects that the input satisfies certain conditions, it will \emph{"flush"} (reset) its states and feed them to a word-level network. Similar operations have also been realized by a gating module on question answering tasks \cite{ke_focused_2018}.}

\vspace{-1ex}
\subsubsection{Temporally Early Exiting in RNNs}
\vspace{-0.25ex}
\label{temporal_early_exit}
Despite that the dynamic RNNs in Sec. \ref{skimming_text} are able to update their states with data-dependent computational costs at each step, all the tokens still must be read, leading to inefficiency in scenarios where the task-relevant results can be obtained before reading the entire sequence.

Ideally, an efficient model should adaptively stop reading before the last step $T$ in Eq. \ref{eq_rnn} is reached, once the captured information is satisfactory to solve the task. For instance, reasoning network (ReasoNet) \cite{shen_reasonet_2017} terminates its reading procedure when sufficient evidence has been found for question answering. 
Similarly, early stopping is implemented for sentence-level \cite{huang_length_2017} and paragraph-level \cite{liu_finding_2020} text classification, respectively. Note that the approaches discussed here focus on making early predictions with respect to the \emph{temporal} dimension of sequential input, rather than along the \emph{depth} dimension of networks as in Sec. \ref{dynamic_depth}.

\vspace{-1ex}
\subsubsection{Jumping in Texts}
\vspace{-0.25ex}
\label{text_jumping}
Although early exiting in Sec. \ref{temporal_early_exit} can largely reduce redundant computation, all the tokens must still be fed to the model one by one. More aggressively, dynamic RNNs could further learn to decide \emph{"where to read"} by strategically skipping some tokens without reading them, and directly jumping to an arbitrary temporal location (see \figurename~\ref{fig_temporal_skim} (d)).

Such dynamic jumping, together with early exiting, is realized in \cite{yu_learning_2017} and \cite{yu_fast_2018}. Specifically, LSTM-Jump \cite{yu_learning_2017} implements an auxiliary unit to predict the jumping stride within a defined range, and the reading process ends when the unit outputs zero. The model in \cite{yu_fast_2018} first decides whether to stop at each step. If not, it will further choose to re-read the current input, or to skip a flexible number of words. Moreover, structural information is exploited by Structural-Jump-LSTM \cite{hansen_neural_2019}, which utilizes an agent to decide whether to jump to the next punctuation. Apart from looking ahead, LSTM-Shuttle \cite{fu_speed_2018} also allows backward jumping to supplement the missed history information.

\vspace{-1.5ex}
\subsection{Temporal-wise Dynamic Video Recognition}
\label{sec:tempoal_video}
\vspace{-0.25ex}
For video recognition, where a video could be seen as a sequential input of frames, temporal-wise dynamic networks are designed to allocate adaptive computational resources for different frames. This can generally be achieved by two approaches: 1) dynamically updating the hidden states in each time step of \emph{recurrent} models (Sec. \ref{recurrent_video}), and 2) performing adaptive \emph{pre-sampling} for key frames (Sec. \ref{frame_sampling}).



\vspace{-1ex}
\subsubsection{Video Recognition with Dynamic RNNs} \label{recurrent_video}
\vspace{-0.25ex}
Video recognition is often conducted via a recurrent procedure, where the video frames are first encoded by a 2D CNN, and the obtained frame features are fed to an RNN sequentially for updating its hidden state. Similar to the approaches introduced in Sec. \ref{temporal_text}, RNN-based adaptive video recognition is typically realized by 1) treating unimportant frames with relatively cheap computation (\emph{"glimpse"}) \cite{wu_liteeval_2019,vaudaux-ruth_actionspotter_2020}; 2) \emph{early exiting} \cite{fan_watching_2018,wu_dynamic_2020}; and 3) performing dynamic \emph{jumping} to decide {"where to see"} \cite{yeung_end--end_2016,su_leaving_2016,fan_watching_2018,wu_adaframe_2019}.


\noindent\textbf{1) Dynamic update of hidden states.} To reduce redundant computation at each time step, LiteEval \cite{wu_liteeval_2019} makes a choice between two LSTMs with different computational costs. ActionSpotter \cite{vaudaux-ruth_actionspotter_2020} decides whether to update the hidden state according to each input frame. {AdaFuse \cite{meng2021adafuse} selectively reuses certain feature channels from the previous step to efficiently make use of historical information.} Recent work has also proposed to adaptively decide the numerical precision \cite{sun2021dynamic} or modalities \cite{weng2021hms,panda2021adamml} when processing the sequential input frames. Such a \emph{glimpse} procedure (i.e. allocating cheap operations to unimportant frames) is similar to the aforementioned text \emph{skimming} \cite{campos_skip_2018,hansen_neural_2019}.

\noindent\textbf{2) Temporally early exiting.} Humans are able to comprehend the contents easily before watching an entire video. Such early stopping is also implemented in dynamic networks to make predictions only based on a portion of video frames \cite{fan_watching_2018,wu_dynamic_2020,ghodrati2021frameexit}. Together with the \emph{temporal} dimension, the model in \cite{wu_dynamic_2020} further achieves early exiting from the aspect of network \emph{depth} as discussed in Sec. \ref{dynamic_depth}.

\noindent\textbf{3) Jumping in videos.} Considering encoding those unimportant frames with a CNN still requires considerable computation, a more efficient solution could be dynamically skipping some frames without watching them. Existing arts \cite{yeung_end--end_2016,su_leaving_2016,alwassel_action_2018} typically learn to predict the location that the network should jump to at each time step.
Furthermore, both early stopping and dynamic jumping are allowed in \cite{fan_watching_2018}, where the jumping stride is limited in a discrete range.
Adaptive frame (AdaFrame) \cite{wu_adaframe_2019} generates a continuous scalar within the range of $[0,1]$ as the relative location. 

\vspace{-1ex}
\subsubsection{Dynamic Key Frame Sampling} \label{frame_sampling}
\vspace{-0.25ex}
Rather than processing video frames recurrently as in Sec. \ref{recurrent_video}, another line of work first performs an adaptive \emph{pre-sampling} procedure, and then makes prediction by processing the selected subset of key frames or clips.

\noindent{\textbf{1) Temporal attention} is a common technique for networks to focus on salient frames.} For face recognition, neural aggregation network \cite{yang_neural_2017} uses \emph{soft} attention to adaptively aggregate frame features. To improve the inference efficiency, \emph{hard} attention is realized to remove unimportant frames iteratively with RL for efficient video face verification \cite{rao_attention-aware_2017}.

\noindent{\textbf{2) Sampling module} is also a prevalent option for dynamically selecting the key frames/clips in a video.} For example, the frames are first sampled uniformly in \cite{tang_deep_2018, wu_multi-agent_2019}, and discrete decisions are made for each selected frame to go forward or backward step by step. As for clip-level sampling, SCSample \cite{korbar_scsampler_2019} is designed based on a trained classifier to find the most informative clips for prediction. Moreover, dynamic sampling network (DSN) \cite{zheng_dynamic_2020} segments each video into multiple sections, and a sampling module with shared weights across the sections is exploited to sample one clip from each section.

{Adjusting multiple factors of deep models simultaneously has attracted researches in both static \cite{han2020model,fan2020rubiksnet} and dynamic networks \cite{li20202d,meng2020ar,wang2021adaptive,pan2021va}. For example, together with \emph{temporal-wise} frame sampling, \emph{spatially} adaptive computation can be achieved by spatial \cite{meng2020ar}/temporal \cite{fayyaz20213d} resolution adaptation and patch selection \cite{wang2021adaptive,verelst2021blockcopy}. It would be promising to exploit the redundancy in both \emph{input data} and \emph{network structure} for further improving the efficiency of deep networks.}

\vspace{-2ex}
\section{Inference and Training}
\label{inference_and_train}
\vspace{-0.25ex}
In previous sections, we have reviewed three different types of dynamic networks (sample-wise (Sec. \ref{sec_sample_wise}), spatial-wise (Sec. \ref{sec_spatially_adaptive}) and temporal-wise (Sec. \ref{sec_temporal_adaptive})). It can be observed that making data-dependent \emph{decisions} at the inference stage is essential to achieve high efficiency and effectiveness. Moreover, \emph{training} dynamic models is usually more challenging than optimizing static networks.

Note that since parameter adaptation (Sec. \ref{adaptive_params}) could be conveniently achieved by differentiable operations, models with dynamic parameters \cite{yang2019condconv,ma_weightnet_2020,hu2018squeeze} can be directly trained by stochastic gradient descent (SGD) without specific techniques. Therefore, in this section we mainly focus on discrete decision making (Sec. \ref{inference}) and its training strategies (Sec. \ref{training}), which are absent in most static models.

\vspace{-1.5ex}
\subsection{Decision Making of Dynamic Networks}
\label{inference}
\vspace{-0.5ex}
As described above, dynamic networks are capable of making data-dependent decisions during inference to transform their architectures, parameters, or to select salient spatial/temporal locations in the input. Here we summarize three commonly seen decision making schemes as follows.

\vspace{-1ex}
\subsubsection{Confidence-based Criteria}
\vspace{-0.25ex}
Many dynamic networks \cite{teerapittayanon2016branchynet,huang2017multi,yang_resolution_2020} are able to output "easy" samples at early exits if a certain confidence-based criterion is satisfied. These methods generally require estimating the confidence of intermediate predictions, which is compared to a predefined threshold for decision making.
In classification tasks, the confidence is usually represented by the maximum element of the \emph{SoftMax} output \cite{huang2017multi,yang_resolution_2020}. Alternative criteria include the entropy \cite{teerapittayanon2016branchynet,xin_deebert_2020} and the score margin \cite{park2015big}. On NLP tasks, a \emph{model patience} is proposed in \cite{zhou_bert_2020}: when the predictions for one sample stay unchanged after a number of classifiers, the inference procedure stops. 

In addition, the halting score in \cite{graves2016adaptive,figurnov2017spatially,dehghani_universal_2019,elbayad_depth-adaptive_2020} could also be viewed as confidence for whether the current feature could be output to the next time step or calculation stage.

Empirically, the confidence-based criteria are easy to implement, and generally require no specific training techniques. A trade-off between accuracy and efficiency is controlled by manipulating the thresholds, which are usually tuned on a validation dataset. Note that the \emph{overconfidence} issue in deep models \cite{guo2017calibration,hein2019relu} might affect the effectiveness of such decision paradigm, when the incorrectly classified samples could obtain a high confidence at early exits.
\vspace{-1ex}
\subsubsection{Policy Networks} \label{inference_policy}
\vspace{-0.25ex}
It is a common option to build an additional policy network learning to adapt the network topology based on different samples. 
Specifically, each input sample is first processed by the policy network, whose output directly determines which parts of the main network should be activated. For example, BlockDrop \cite{wu2018blockdrop} and GaterNet \cite{chen2019you} use a policy network to adaptively decide the \emph{depth} and \emph{width} of a backbone network. More generally, dynamic routing in a \emph{SuperNet} can also be controlled by a policy network \cite{cheng2020instanas}.

One possible limitation of this scheme is that the architectures and the training process of some policy networks are developed for a specific backbone \cite{wu2018blockdrop,chen2019you}, and may not be easily adapted to different architectures. 


\vspace{-1ex}
\subsubsection{Gating Functions} \label{inference_gating}
\vspace{-0.25ex}
Gating function is a general and flexible approach to decision making in dynamic networks. It can be conveniently adopted as a plug-in module at arbitrary locations in any backbone network.
During inference, each module is responsible for controlling the local inference graph of a layer or block. The gating functions take in intermediate features and efficiently produce binary-valued gate vectors to decide: 1) which channels need to be activated \cite{lin2017runtime,gao2018dynamic, herrmann2018end,bejnordi2019batch,chen2019self} \emph{width}, 2) which layers need to be skipped \cite{veit2018convolutional,wang2018skipnet,wang_dual_2020,xia2020fully}, 3) which paths should be selected in a SuperNet \cite{li_learning_2020}, or 4) what locations of the input should be allocated computations \cite{kong_pixel-wise_2019,verelst_dynamic_2020,xie_spatially_2020, meng2021adafuse}.

Compared to the aforementioned decision policies, the gating functions demonstrate notable generality and applicability. However, due to their lack of differentiability, these gating functions usually need specific training techniques, which will be introduced in the following Sec. \ref{training}.


\vspace{-2ex}
\subsection{Training of Dynamic Networks}
\vspace{-0.5ex}
\label{training}
Besides architecture design, training is also essential for dynamic networks. Here we summarize the existing training strategies for dynamic models from the perspectives of objectives and optimization.
\begin{table*}
  \scriptsize
  \vspace{-2ex}
  \caption{Applications of Dynamic Networks. For the type column, Sa, Sp and Te stand for sample-wise, spatial-wise and temporal-wise respectively.}
  \label{tab_tasks}
  \begin{center}
    \vspace{-6ex}
    \resizebox{\linewidth}{!}{
    \begin{tabular}{cccc}
      \toprule
      \textbf{Fields} & \textbf{Data} & \textbf{Type} & \textbf{Subfields \& references} \\
      \midrule
      & \multirow{6}*{\textbf{Image}} & \multirow{2}*{Sa} & Object detection (face \cite{rowley1998neural,viola_robust_2004,li_convolutional_2015}, facial point \cite{sun2013deep}, pedestrian \cite{angelova_real_time_2015}, general \cite{yang_exploit_2016,figurnov2017spatially,zhou_adaptive_2017,yang_metaanchor_2018,chen_adaptive_2019}) \\
      & & & Image segmentation \cite{tokunaga2019adaptive, li_learning_2020,wang2020deep}, Super resolution \cite{riegler_conditioned_2015}, Style transfer \cite{shen_neural_2018},  Coarse-to-fine classification \cite{jiang2020learning}\\ 
      \cmidrule{3-4}
      & &  \multirow{3}*{Sa \& Sp} & Image segmentation \cite{li_not_2017,kong_pixel-wise_2019,xie_spatially_2020,kirillov2020pointrend,wang_carafe_2019,he_dynamic_2019,wang_adaptively_2019,marin_efficient_2019,li_dense_2017,roy2018concurrent,wu_dynamicAttention_2020,zhong_squeeze-and-attention_2020}, Image-to-image \\
      & & & translation \cite{huang2018multimodal}, Object detection \cite{hao_scale-aware_2017, verelst_dynamic_2020,xie_spatially_2020,dai2017deformable, zhu_deformable_2019},  Semantic image synthesis \cite{liu_learning_2019,park_semantic_2019, zhu_sean_2020}, \\
      \multirow{1}*{\textbf{Computer}} & & & Image denoising \cite{chang2020spatial}, Fine-grained classification \cite{xiao_application_2015,zheng_learning_2017,fu_look_2017,recasens_learning_2018} Eye tracking \cite{recasens_learning_2018}, Super resolution \cite{bhowmik_training-free_2017,hu_meta-sr_2019,sun_learned_2020} \\
      \cmidrule{3-4}
      \multirow{1}*{\textbf{Vision}}& &  Sa \& Sp \& Te & General classification \cite{mnih_recurrent_2014,rosenfeld_visual_2016,wang2020glance}, Multi-object classification \cite{ba_multiple_2015,eslami_attend_2016}, Fine-grained classification \cite{li_dynamic_2017} \\
      \cmidrule{2-4}
      & \multirow{4}*{\textbf{Video}} & Sa & Multi-task learning (human action recognition and frame prediction) \cite{diba_dynamonet_2019} \\
      \cmidrule{3-4}
      & &  \multirow{2}*{Sa \& Te} & Classification (action recognition) \cite{fan_watching_2018,wu_adaframe_2019,gao_listen_2020,wu_liteeval_2019,tang_deep_2018,korbar_scsampler_2019,wu_multi-agent_2019,zheng_dynamic_2020,meng2020ar}, Semantic segmentation \cite{xu2018dynamic}\\
      & & & Video face recognition \cite{yang_neural_2017,rao_attention-aware_2017}, Action detection \cite{yeung_end--end_2016,su_leaving_2016}, Action spotting \cite{alwassel_action_2018,vaudaux-ruth_actionspotter_2020} \\
      \cmidrule{3-4}
      & &  Sa \& Sp \& Te & Classification \cite{meng2020ar,wang2021adaptive}, Frame interpolation \cite{niklaus_video_2017,niklaus_video_2017-1}, Super resolution \cite{jo_deep_2018}, Video deblurring \cite{hyun2017online,zhou_spatio-temporal_2019}, Action prediction \cite{chen_part-activated_2018}\\
      \cmidrule{2-4}
      & \textbf{Point Cloud} & Sa \& Sp  & 3D Shape classification and segmentation, 3D scene segmentation \cite{thomas_kpconv_2019}, 3D semantic scene completion \cite{li2020anisotropic} \\
      \midrule
      \multirow{2}*{\textbf{Natural}} & \multirow{3}*{\textbf{Text}} & Sa  & Neural language inference, Text classification, Paraphrase similarity matching, and Sentiment analysis \cite{schwartz_right_2020,zhou_bert_2020} \\
      \cmidrule{3-4}
      \textbf{Language} & &  \multirow{2}*{Sa \& Te} & Language modeling \cite{graves2016adaptive,shazeer2017outrageously, ha2016hypernetworks,jernite_variable_2017,chung_hierarchical_2017}, Machine translation \cite{shazeer2017outrageously,dehghani_universal_2019,elbayad_depth-adaptive_2020}, Classification \cite{huang_length_2017,yu_fast_2018,liu_finding_2020}, \\
      \textbf{Processing} & & &  Sentiment analysis \cite{hansen_neural_2019,tao_skipping_2019,seo_neural_2018,yu_learning_2017,fu_speed_2018}, Question answering \cite{dehghani_universal_2019,hansen_neural_2019,seo_neural_2018,ke_focused_2018,shen_reasonet_2017} \\
      \midrule
      \textbf{Cross-Field} & \multicolumn{3}{c}{Image captioning \cite{xu_show_2015, chen_sca-cnn_2017}, {Video captioning \cite{hori2017attention,sun2019videobert}, Visual question answering \cite{gao_question-guided_2018, de2017modulating,perez2018film}, Multi-modal sentiment analysis \cite{zadeh2018multimodal,rahman2020integrating}}}  \\
      \midrule
      \multirow{2}{*}{\textbf{Others}} &  \multicolumn{3}{c}{{Time series forecasting} \cite{cinar2017position,fan2019multi,jin2021inter}, Link prediction \cite{jiang_adaptive_2019}, {Recommendation system} \cite{ma2018modeling,song2019session,song2019autoint,huang2020efficient}}\\
      & \multicolumn{3}{c}{Graph classification \cite{simonovsky_dynamic_2017}, {Document classification} \cite{wang_adaptively_2019,nikolentzos2020message,choi2020improving,zhang2020text}, Stereo confidence estimation \cite{kim_laf-net_2019}}\\
      
  \bottomrule
    \end{tabular}}
  \end{center}
  \vskip -0.3in
\end{table*}
\vspace{-1ex}
\subsubsection{Training Objectives for Efficient Inference}\label{sec_train_objectives}
\vspace{-0.25ex}

\noindent\textbf{1) Training multi-exit networks.} We first notice that early-exiting dynamic networks \cite{huang2017multi,yang_resolution_2020} are generally trained by minimizing a weighted cumulative loss of intermediate classifiers. One challenge for training such models is the joint optimization of multiple classifiers, which may interfere with each other. MSDNet \cite{huang2017multi} alleviates the problem through its special architecture design. Several improved training techniques \cite{li2019improved} are proposed for multi-exit networks, including a gradient equilibrium algorithm to stable the training process, and a bi-directional knowledge transfer approach to boost the collaboration of classifiers. {For temporal-wise early exiting, the training of the policy network in FrameExit \cite{ghodrati2021frameexit} is supervised by pseudo labels.}

\noindent\textbf{2) Encouraging sparsity.} Many dynamic networks adapt their inference procedure by conditionally activating their computational units \cite{wang2018skipnet,bejnordi2019batch} or strategically sampling locations from the input \cite{xie_spatially_2020}.
Training these models without additional constraints would result in superfluous computational redundancy, as a network could tend to activate all the candidate units for minimizing the task-specific loss. 

The overall objective function for restraining such redundancy are typically written as $\mathfrak{L}\! =\!\mathfrak{L}_\mathrm{task}\!+\!\gamma \mathfrak{L}_\mathrm{sparse}$, where $\gamma$ is the hyper-parameter balancing the two items for the trade-off between accuracy and efficiency. In real-world applications, the second item can be designed based on the gate/mask values of candidate units (e.g. channels \cite{bejnordi2019batch,herrmann2018end}, layers \cite{wang2018skipnet,veit2018convolutional} or spatial locations \cite{xie_spatially_2020}). Specifically, one may set a target activation rate \cite{veit2018convolutional,herrmann2018end} or minimizing the $\mathcal{L}_1$ norm of the gates/masks \cite{xie_spatially_2020}. It is also practical to directly optimize a resource-aware loss (e.g. FLOPs) \cite{wang_dual_2020,li_learning_2020,verelst_dynamic_2020}, which can be estimated according to the input and output feature dimension for every candidate unit.

\noindent{\textbf{3) Others.} Note that extra loss items are mostly designed for but not limited to improving efficiency. Take \cite{fu_look_2017} as an example, the model progressively focuses on a selected region, and is trained with an additional \emph{inter-scale pairwise ranking loss} for proposing more discriminative regions. Moreover, knowledge distilling is utilized to boost the co-training of multiple sub-networks in \cite{hua2019channel} and \cite{li2019improved}.}

\vspace{-1ex}
\subsubsection{Optimization of Non-differentiable Functions}
\vspace{-0.25ex}
A variety of dynamic networks contain non-differentiable functions that make discrete decisions to modify their architectures or sampling spatial/temporal locations from the input. These functions can not be trained directly with back-propagation. Therefore, specific techniques are studied to enable the end-to-end training as follows.

\noindent\textbf{1) Gradient estimation} is proposed to approximate the gradients for those non-differentiable functions and enable back-propagation. In \cite{bengio2013estimating,chung_hierarchical_2017}, straight-through estimator (STE) is exploited to heuristically copy the gradient with respect to the stochastic output directly as an estimator of the gradient with respect to the \emph{Sigmoid} argument.

\noindent\textbf{2) Reparameterization} is also a popular technique to optimize the discrete decision functions. For instance, the gating functions controlling the network width \cite{herrmann2018end} or depth \cite{veit2018convolutional} can both be trained with \emph{Gumbel SoftMax} \cite{gumbel1954statistical, jang2016categorical}, which is also used for pixel-level dynamic convolution \cite{xie_spatially_2020,verelst_dynamic_2020}.  An alternative technique is \emph{Improved SemHash} \cite{kaiser2018discrete} adopted in \cite{chen2019self} and \cite{chen2019you} to train their hard gating modules.

{Note that although these reparameterization techniques enable joint optimizing dynamic models together with gating modules in an end-to-end fashion, they usually lead to a longer training process for the decision functions to converge into a stable situation \cite{dong_more_2017}. Moreover, the model performance might be sensitive to some extra hyper-parameters (e.g. temperature in \emph{Gumbel SoftMax}), which might also increase the training cost for these dynamic networks.}

\noindent\textbf{3) Reinforcement learning (RL)} is widely exploited for training non-differentiable decision functions. In specific, the backbones are trained by standard SGD, while the agents (either policy networks in Sec. \ref{inference_policy} or gating functions in Sec. \ref{inference_gating}) are trained with RL to take discrete actions for dynamic inference graphs \cite{lin2017runtime,wang2018skipnet,wu2018blockdrop} or spatial/temporal sampling strategies \cite{wang2020glance,wu_multi-agent_2019}.

{One challenge for RL-based training is the design of reward functions, which is important to the accuracy-efficiency tradeoff of dynamic models. Commonly seen reward signals are usually constructed to minimize a penalty item of the computational cost \cite{lin2017runtime,wang2018skipnet}. Moreover, the training could be costly due to a multi-stage procedure: a pre-training process may be required for the backbone networks before the optimization of decision \cite{wu2018blockdrop} or sampling \cite{wang2020glance} modules, and joint finetuning may be indispensable finally.}

\vspace{-2ex}
\section{Application of Dynamic Networks}\label{sec_tasks}
\vspace{-0.25ex}
In this section, we summarize the applications of dynamic networks. Representative methods are listed in Table \ref{tab_tasks} based on the input data modality.

For image recognition, most dynamic CNNs are designed to conduct \emph{sample-wise} or \emph{spatial-wise} adaptive inference on classification tasks, and many inference paradigms can be generalized to other applications. Note that as mentioned in Sec. \ref{region_level}, the object recognition could be formulated as a sequential decision problem \cite{li_dynamic_2017,wang2020glance}. By allowing early exiting in these approaches, \emph{temporally} adaptive inference procedure could also be enabled.

For text data, reducing its intrinsic temporal redundancy has attracted great research interests, and the inference paradigm of \emph{temporal-wise} dynamic RNNs (see Sec. \ref{temporal_text}) is also general enough to process audios \cite{tavarone_conditional-computation-based_2018}. Based on large language models such as Transformer \cite{vaswani2017attention} and BERT \cite{devlin_bert_2019}, adaptive depths \cite{liu_fastbert_2020,xin_deebert_2020,schwartz_right_2020,zhou_bert_2020} are extensively studied to reduce redundant computation in network architectures.

For video-related tasks, the three types of dynamic inference can be implemented simultaneously \cite{li_dynamic_2017,niklaus_video_2017,niklaus_video_2017-1,wang2021adaptive}. However, for the networks that do not process videos recurrently, e.g. 3D CNNs \cite{tran2015learning, carreira2017quo,he2019stnet}, most of them still follow a static inference scheme. Few researches have been committed to building dynamic 3D CNNs \cite{li20202d}, which might be an interesting future research direction.


{Moreover, dynamic networks (especially the attention mechanism) have also been applied to dynamically fuse the features from different modalities in some multi-modal learning tasks, e.g. RGB-D image segmentation \cite{wang2020deep} and image/video captioning \cite{xu_show_2015, chen_sca-cnn_2017,hori2017attention,sun2019videobert}.}


Finally, dynamic networks have also been exploited to tackle some fundamental problems in deep learning. For example, multi-exit models can be used to: 1) alleviate the \emph{over-thinking} issue while reducing the overall computation \cite{wang2017idk,kaya_shallow-deep_2019}; 2) perform \emph{long-tailed classification} \cite{duggal_elf_2020} by inducing early exiting in the training stage; and 3) improve the model \emph{robustness} \cite{hu_triple_2020}. For another example, the idea of dynamic routing is implemented for: 1) {reducing the \emph{training cost} under a multi-task setting \cite{rosenbaum_routing_2018}} and 2) finding the optimal fine-tuning strategy for per example in \emph{transfer learning} \cite{Guo_2019_CVPR_spottune}.

\vspace{-2ex}
\section{{Challenges and Future Directions}}
\label{sec:discussion}
\vspace{-0.5ex}
{Though recent years have witnessed significant progress in the research of dynamic neural networks, there still exist many open problems that are worth exploring.} In this section, we summarize a few challenges together with possible future directions in this field.

\vspace{-1ex}
\subsection{Theories for Dynamic Networks}
\vspace{-0.5ex}
Despite the success of dynamic neural networks, relatively few researches has been committed to analyze them from the theoretical perspective. In fact, theories for a deep understanding of current dynamic learning models and further improving them in principled ways are highly valuable. {Notably, it has been proven that a dynamic network with an adaptive width can preserve the representation power of an unsparsified model \cite{pmlr-v115-wang20d}.} However, there are more theoretical problems that are fundamental for dynamic networks. Here we list several of them as follows.

\noindent\textbf{1) Optimal decision in dynamic networks.}
An essential operation in most dynamic networks (especially those designed for improving computational efficiency) is making data-dependent decisions, e.g., determining whether a module should be evaluated or skipped. Existing solutions either use confidence-based criteria, or introduce policy networks and gating functions. Although being effective in practice (as mentioned in Sec. \ref{inference_and_train}), they may not be optimal and lack theoretical justifications. Take early exiting as an example, the current heuristic methods \cite{huang2017multi,yang_resolution_2020} might face the issues of overconfidence, high sensitivity for threshold setting and poor transferability. As for policy networks or gating modules, runtime decisions can be made based on a learned function.
However, they often introduce extra computations, and usually require a long and unstable training procedure. Therefore, principled approaches with theoretical guarantees for decision function design in dynamic networks is a valuable research topic.

\noindent\textbf{2) Generalization issues.} 
In a dynamic model, a sub-network might be activated for a set of test samples that are not uniformly sampled from the data distribution, e.g., smaller sub-networks tend to handle ``easy'' samples, while larger sub-networks are used for ``hard'' inputs \cite{huang2017multi}. This brings a divergence between the training data distribution and that of the inference stage, and thus violates the common \emph{i.i.d.} assumption in classical machine learning. Therefore, it would be interesting to develop new theories to analyze the generalization properties of dynamic networks under such distribution mismatch. Note that transfer learning also aims to address the issue of distributional shift at test time, but the samples of the target domain are assumed to be accessible in advance. In contrast, for dynamic models, the test distribution is not available until the training process is finished, when the network architecture and parameters are finalized. This poses greater challenges than analyzing the generalization issues in transfer learning.

\subsection{Architecture Design for Dynamic Networks}
Architecture design has been proven to be essential for deep networks. Existing researches on architectural innovations are mainly proposed for static models \cite{he2016deep,huang2017densely,howard2017mobilenets}, while relatively few are dedicated to developing architectures specially for dynamic networks. It is expected that architectures developed specifically for dynamic networks may further improve their effectiveness and efficiency. For example, the interference among multiple classifiers in an early-exiting network could be mitigated by a carefully designed multi-scale architecture with dense connections \cite{huang2017multi}. 

Possible research direction include designing dynamic network structures either by hand (as in \cite{huang2017multi,yang_resolution_2020,Guo_2019_CVPR,dehghani_universal_2019}), or by leveraging the NAS techniques (as in \cite{yuan2019s2dnas,cheng2020instanas}). Moreover, considering the popularity of Transformers \cite{dosovitskiy2020image}, {recent work has proposed dynamic vision Transformers with adaptive early exiting \cite{wang2021images} or token sparsification \cite{rao2021dynamicvit,pan2021ia}.} Developing a dynamic version of this family of models could also be an interesting direction.

{Note that the research on dynamic networks differs from a seemingly close topic, i.e. model compression \cite{huang2018condensenet,hubara2016binarized,low_rank}. One common goal of them is improving the network efficiency with minimal accuracy drop. However, model compression may focus on reducing the \emph{size} of deep networks, while dynamic networks pay more attention to the \emph{computation}, even at the price of slightly \emph{increasing} model size \cite{wang2018skipnet,lin2017runtime}. Moreover, model compression typically adopts pruning \cite{huang2018condensenet} or quantization \cite{hubara2016binarized} techniques to produce compact \emph{static} models, which treat all the inputs in the same way. In contrast, dynamic networks perform data-dependent computation on different inputs, which can effectively reduce the intrinsic redundancy in static models.}
\vspace{-1ex}
\subsection{Applicability for More Diverse Tasks}
\vspace{-0.5ex}
Many existing dynamic networks (e.g., most of the sample-wise adaptive networks) are designed specially for classification tasks, and cannot be applied to other vision tasks such as object detection and semantic segmentation. The difficulty arises from the fact that for these tasks there is no simple criterion to assert whether an input image is easy or hard, as it usually contains multiple objects and pixels that have different levels of difficulty. Although many efforts, e.g., spatially adaptive models \cite{figurnov2017spatially,xie_spatially_2020,wang2020glance} and soft attention based models \cite{hu2018squeeze,woo_cbam_2018,yang2019condconv}, have been made to address this issue, it remains a challenging problem to develop a unified and elegant dynamic network that can serve as an off-the-shelf backbone for a variety of tasks.

\vspace{-1.5ex}
\subsection{Gap between Theoretical \& Practical Efficiency} \label{discuss_implement}
\vspace{-0.5ex}
The current deep learning hardware and libraries are mostly optimized for static models, and they may not be friendly to dynamic networks. Therefore, we usually observe that the practical runtime of dynamic models lags behind the theoretical efficiency. For example, some spatially adaptive networks involve sparse computation, {which is known to be inefficient on modern computing devices due to the memory access bottleneck \cite{xie_spatially_2020}. A recent line of work focuses on the codesign of algorithm and hardware for accelerating deep models on platforms with more flexibility such as FPGA \cite{yang2019synetgy}. Many input-dependent operations, including pixel-level dynamic computation \cite{albericio_cnvlutin_2016,lin2017predictivenet,huang2021codenet}, adaptive channel pruning \cite{akhlaghi2018snapea,hua2019boosting} and early exiting \cite{paul2019hardware}, have also been tailored together with hardware for further improving their practical efficiency. It is an interesting research direction to simultaneously optimize the algorithm, hardware and deep learning libraries to harvest the theoretical efficiency gains of dynamic networks.}

{In addition, a data-dependent inference procedure, especially for the dynamic \emph{architectures}, usually requires a model to handle input samples sequentially, which also poses challenge for parallel computation. Although inference with batches has been enabled for early-exiting networks \cite{wang2021images}, the conflict between adaptive computational graph and parallel computation still exists for other types of dynamic architectures. This issue is mitigated in the scenario of mobile/edge computing, where the input signal by itself is sequential and the computing hardware is less powerful than high-end platforms. However, designing dynamic networks that are more compatible with existing hardware and software is still a valuable and challenging topic.}

\vspace{-1.5ex}
\subsection{Robustness Against Adversarial Attack}
\vspace{-0.5ex}
Dynamic models may provide new perspectives for the research on adversarial robustness of deep neural networks. {For example, recent work \cite{hu_triple_2020} has leveraged the multi-exit structure to improve the robustness against adversarial attacks. Moreover, traditional attacks are usually aimed at causing \emph{misclassification}. For dynamic networks, it is possible to launch attacks on \emph{efficiency} \cite{haque2020ilfo,hong2020panda}. Specifically, by adjusting the objective function of the adversarial attack, input-adaptive models could be fooled to activate all their intermediate layers \cite{haque2020ilfo} or yielding confusing predictions at early exits \cite{hong2020panda} even for "easy" samples. It has also been observed that the commonly used adversarial training is not effective to defend such attacks.} The robustness of dynamic network is an interesting yet understudied topic.


\vspace{-1.5ex}
\subsection{Interpretability}
\vspace{-0.5ex}
Dynamic networks inherit the black-box nature of deep learning models, and thus also invite research on interpreting their working mechanism. What is special here is that the adaptive inference paradigm, e.g., spatial/temporal adaptiveness, conforms well with that of the human visual system, and may provide new possibilities for making the model more transparent to humans. In a dynamic network, it is usually convenient to analyze which part of the model is activated for a given input or to locate which part of the input the model mostly relies on in making its prediction. {It is expected that the research on dynamic network will inspire new work on the interpretability of deep learning.}
\vspace{-2ex}



%

\appendices


\ifCLASSOPTIONcompsoc
  \section*{Acknowledgments}
\else
  \section*{Acknowledgment}
\fi

This work is supported in part by the National Science and Technology Major Project of the Ministry of Science and Technology of China under Grants 2018AAA0100701,  the National Natural Science Foundation of China under Grants 61906106 and 62022048, the Institute for Guo Qiang of Tsinghua University and Beijing Academy of Artificial Intelligence.

\ifCLASSOPTIONcaptionsoff
  \newpage
\fi



%

%


\small{
    \bibliography{reference}
}

\end{document}